\documentclass[sigconf,10pt]{acmart}
 
\usepackage{amsmath}
\usepackage{multirow}
\usepackage{tikz}
\usepackage{makecell}
\usepackage{graphicx} 
\usepackage{tikz}
\usepackage{subfigure}
\usepackage{booktabs,lscape}
\usepackage{xspace}
\usepackage{bm}  
\usepackage{url}
\usepackage{siunitx}

\usepackage{algorithm}
\usepackage{algorithmicx}
\usepackage[noend]{algpseudocode}
\usepackage{colortbl}
 
\newcommand{\systemname}{{\sf Map++}\xspace}
\newcommand{\dataset}{{Future-City}\xspace}
\newcommand{\datasetgarage}{{Indoor-Garage}\xspace}
\newcommand{\datasetbuilding}{{Outdoor-Plaza}\xspace}

\definecolor{codecolor}{rgb}{0, 0.8, 0.4}

\AtBeginDocument{%
  }
    
\begin{document}

\acmYear{2024}\copyrightyear{2024}
\setcopyright{acmlicensed}
\acmConference[ACM MobiCom '24]{International Conference On Mobile Computing And Networking}{September 30--October 4, 2024}{Washington D.C., DC, USA}
\acmBooktitle{International Conference On Mobile Computing And Networking (ACM MobiCom '24), September 30--October 4, 2024, Washington D.C., DC, USA}
\acmDOI{10.1145/3636534.3649386}
\acmISBN{979-8-4007-0489-5/24/09}

\title{\systemname: Towards User-Participatory Visual SLAM Systems with Efficient Map Expansion and Sharing}

\renewcommand{\shorttitle}{ \systemname}

\begin{sloppypar}
 \begin{abstract} 
Constructing precise 3D maps is crucial for the development of future map-based systems such as self-driving and navigation.
However, generating these maps in complex environments, such as multi-level parking garages or shopping malls, remains a formidable challenge. 
In this paper, we introduce a participatory sensing approach that delegates map-building tasks to map users, thereby enabling cost-effective and continuous data collection. 
The proposed method harnesses the collective efforts of users, facilitating the expansion and ongoing update of the maps as the environment evolves.

We realized this approach by developing Map++, an efficient system that functions as a plug-and-play extension, supporting participatory map-building based on existing SLAM algorithms. 
Map++ addresses a plethora of scalability issues in this participatory map-building system by proposing a set of lightweight, application-layer protocols.
We evaluated \systemname in four representative settings: an indoor garage, an outdoor plaza, a public SLAM benchmark, and a simulated environment. The results demonstrate that \systemname can reduce traffic volume by approximately 46\%  with negligible degradation in mapping accuracy, i.e., less than 0.03m compared to the baseline system. It can support approximately $2 \times$  as many concurrent users as the baseline under the same network bandwidth. Additionally, for users who travel on already-mapped trajectories, they can directly utilize the existing maps for localization and save 47\% of the CPU usage.
\end{abstract}




\author{Xinran Zhang$^\dagger$, Hanqi Zhu$^\dagger$, Yifan Duan$^\dagger$, Wuyang Zhang$^\ddagger$, Longfei Shangguan$^\S$, Yu Zhang$^{\dagger}$, Jianmin Ji$^{\dagger}$, Yanyong Zhang$^{\dagger}$\\
\small $^\dagger$ University of Science and Technology of China, 
\small $^\ddagger$ Meta, 
\small $^\S$ University of Pittsburgh \\ 
\vspace{-5pt}
}

\renewcommand{\shortauthors}{Zhang et al.}

\maketitle

\vspace{-4pt}
\section{Introduction} \label{sec:intro} \vspace{-2pt}
Exploring and mapping uncharted environments has always been a captivating and enduring challenge, from the earliest human migrations to modern space exploration. Recently, with the advance of robotics and autonomous driving technology, high-resolution 3D maps have received a great deal of attention. Envision the following scenario: as you approach a massive, bustling parking garage unfamiliar to you, just minutes before a crucial meeting, you wish your car to autonomously locate an available parking spot and park itself securely. Given that numerous cars today can self-park (once the parking spot is identified), this aspiration is a reasonable leap forward. To realize this vision, a comprehensive, navigate-able 3D map of the garage is urgently needed.

\begin{figure*}
\centering
\includegraphics[width=.74\linewidth] {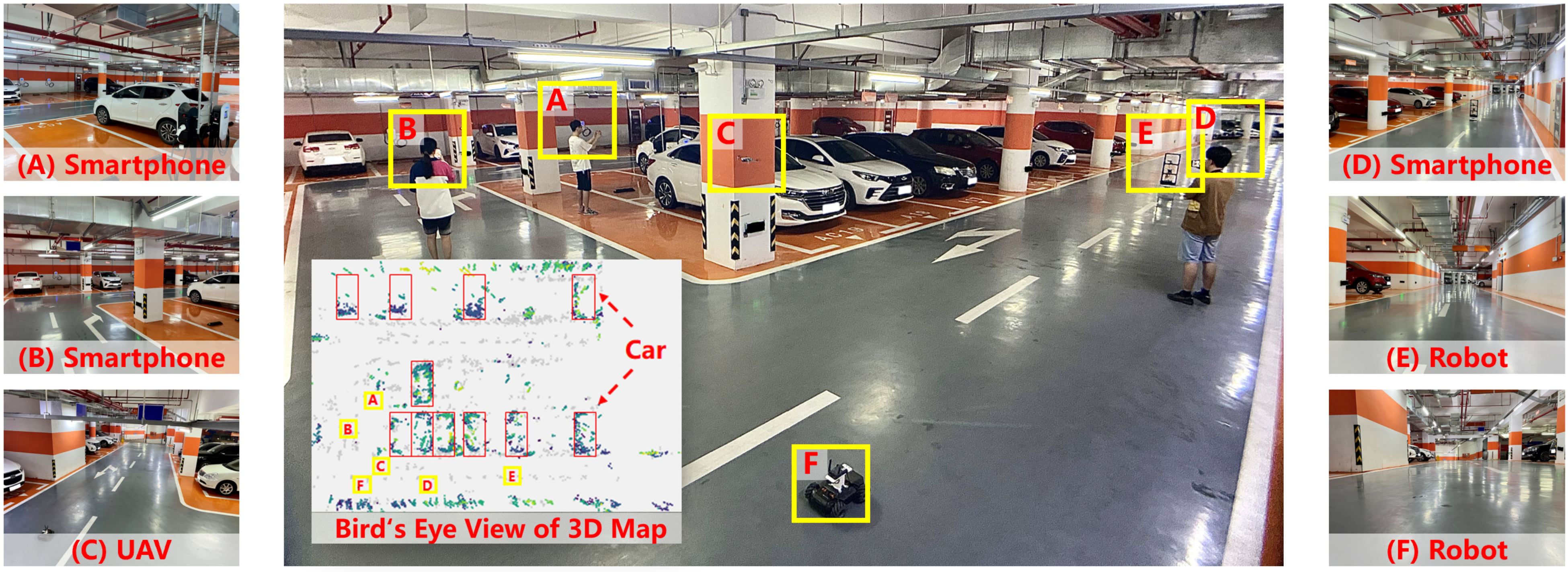} 
\vspace{-11pt}
\caption{\small\label{fig:showcase} A User-Participatory SLAM system. Users upload data to contribute to the map (shown in the bottom left corner of the garage) on the server.}
\vspace{-14pt}
\end{figure*}

The Simultaneous Localization and Mapping (SLAM) technique is crucial to building such 3D maps.
SLAM enables continuous user localization/navigation while simultaneously modeling their environment using data collected from various sensors, such as cameras, depth sensors, LiDARs~\cite{orbslam3, rgbd-slam}, which can be attached to the user (e.g., the car, the smartphone) without any infrastructure within the environment. 

Despite extensive research on SLAM algorithms in the robotics domain, practical SLAM systems, particularly those capable of mapping sizable and complex areas such as multi-level parking garages, keeping the map up to date, and maintaining the map service for long periods of operations remain elusive.  
The main challenges in building a functional SLAM system arise from the difficulty of collecting comprehensive and fine-grained sensor data of the area of interest over a long period of time at a low cost~\cite{lajoie2021towards, zou2019collaborative}.

In this work, we propose collaborative \emph{\textbf{user-participatory}} SLAM systems, shown in Fig.~\ref{fig:showcase}, that leverage the widespread availability of onboard cameras on users' mobile devices or cars for gathering map data and constructing a global 3D map at edge/cloud servers. 
Users contribute to map construction in a laissez-faire fashion, not following instructions to move~\cite{capponi2019survey}.
By harnessing the collective efforts of users, it facilitates convenient, low-cost, and continuous data collection, enabling the map to expand as users move through the space and keep updated as the environment changes. In turn, the users can obtain and utilize the up-to-date 3D map on their devices.  
Precisely, the devices capture the surroundings using cameras and upload the data to a map server, which then merges the data into the global map and conducts a global optimization process. 
We believe that this map-building and maintenance approach can provide effective solutions across numerous environments.  The resulting 3D maps can serve as valuable additions to centralized commercial maps such as Google Maps, in terms of both map coverage and the ability to navigate mobile devices and cars.  %

Despite its great promises, user-participatory SLAM faces several fundamental yet intertwined challenges. 
The primary challenge lies in the excessive map data redundancy. 
User-participatory mapping allows users to voluntarily contribute data, and upload it to the server while following their trajectories. When two devices capture the environment at similar locations, either simultaneously or at distinct instances, their sensing ranges are likely to overlap, producing redundant data. 
Given that most users traverse shared roads and paths, there exists a high degree of data redundancy, resulting in significant waste in network bandwidth, processing power, and memory usage.  Additionally, the frequent transmission of map data may discourage user engagement due to limited (or, expensive) resources on users' mobile devices (including cars). In addition to transmitting map data, several SLAM functions need to be performed on user devices, including pose estimation, map data generation,  and local optimization, which can sum up to high resource consumption. 

In this work, we address this challenge by identifying the degree of redundancy between newly acquired map data and the existing global map and only requesting ``fresh'' data to be transmitted to the server. This can tremendously reduce resource consumption on the server as well as the participation cost of users.
To achieve this goal, we devise a lightweight redundancy-checking mechanism utilizing two types of map metadata -- 
the device dispatches, instead of the raw data, its pose (location and orientation) to the server, which then constructs a view cone representing the 3D field of view (FOV) of the camera at the pose. 
By evaluating the overlap between this view cone and the global map through an efficient spatial sampling technique, we can determine the overlap between the new map data and the global map with a minimal cost. 
 
Based on the overlap evaluation outcomes, we determine if the device's current location is previously ``seen'' or new. If the location has been seen and mapped, the user does not need to upload map data, significantly reducing the processing/networking/memory resources. Meanwhile, the server shares the global map's surrounding portion directly with the device. 
Leveraging the shared map, the device skips the expensive local optimization step, conserving computation resources and battery energy consumption. 
Furthermore, the server can suitably enlarge the shared map portion to include the device's future locations, further reducing the system overhead. As such, for the first time, we can provide the 3D map service to passing-by devices, making the users feel more rewarded and worthwhile.    
If the location is new or partially new, the corresponding new map data must be uploaded to expand the global map. 
Towards this goal, we devise a redundancy control method, involving first removing all the map data that are redundant with the global map and then strategically injecting a minimal amount of redundant map data that are frequently observed and can thus be exploited for better map optimization purposes.

In this work, we design \systemname, an efficient user-participatory SLAM system that functions as a plug-and-play extension to support existing SLAM algorithms with minimal resource consumption.
We have implemented \systemname and integrated it with the open-source project of  Covins~\cite{covins} based on ORB-SLAM3~\cite{orbslam3}, a state-of-the-art visual SLAM algorithm. To summarize,
the main contributions of this paper are as follows: \vspace{-3pt}
\begin{enumerate}
     \item We are the first, to the best of our knowledge, to propose a user-participatory SLAM framework that aims to build a shared map with low resource costs by exploiting user trajectory properties. 
     Compared to a trajectory-unaware distributed SLAM system, our system maximizes the number of participating users under given resource constraints while maintaining SLAM accuracy.
     As the map expands, subsequent users can access the map as needed. 
     Also, our system can support efficient map updates without incurring skyrocketing memory costs.
     
    \item To minimize data redundancy for reduced computation and communication costs, we devise a set of protocols and algorithms, including metadata-based overlapping assessment, global map sharing for seen locations, and global map expansion for new locations.
    
    \item  We thoroughly evaluate the system in four distinct settings with heterogeneous cameras, including two real-world scenarios, a public dataset, and a simulated environment. \systemname manages to reduce approximately 46\% traffic volume for map expansion with only a slight degradation in accuracy, i.e., less than 0.03m compared to the baseline system. Consequently,  it can support approximately $2 \times$  as many users as the baseline under the same network bandwidth when they participate in mapping at the same time. Additionally, for users who travel on previously-mapped trajectories, they can directly utilize the existing maps for their operations such as localization and save 47\% of the CPU usage.
\end{enumerate}

\begin{figure}[!t]
\centering
\includegraphics[width=0.83\linewidth] {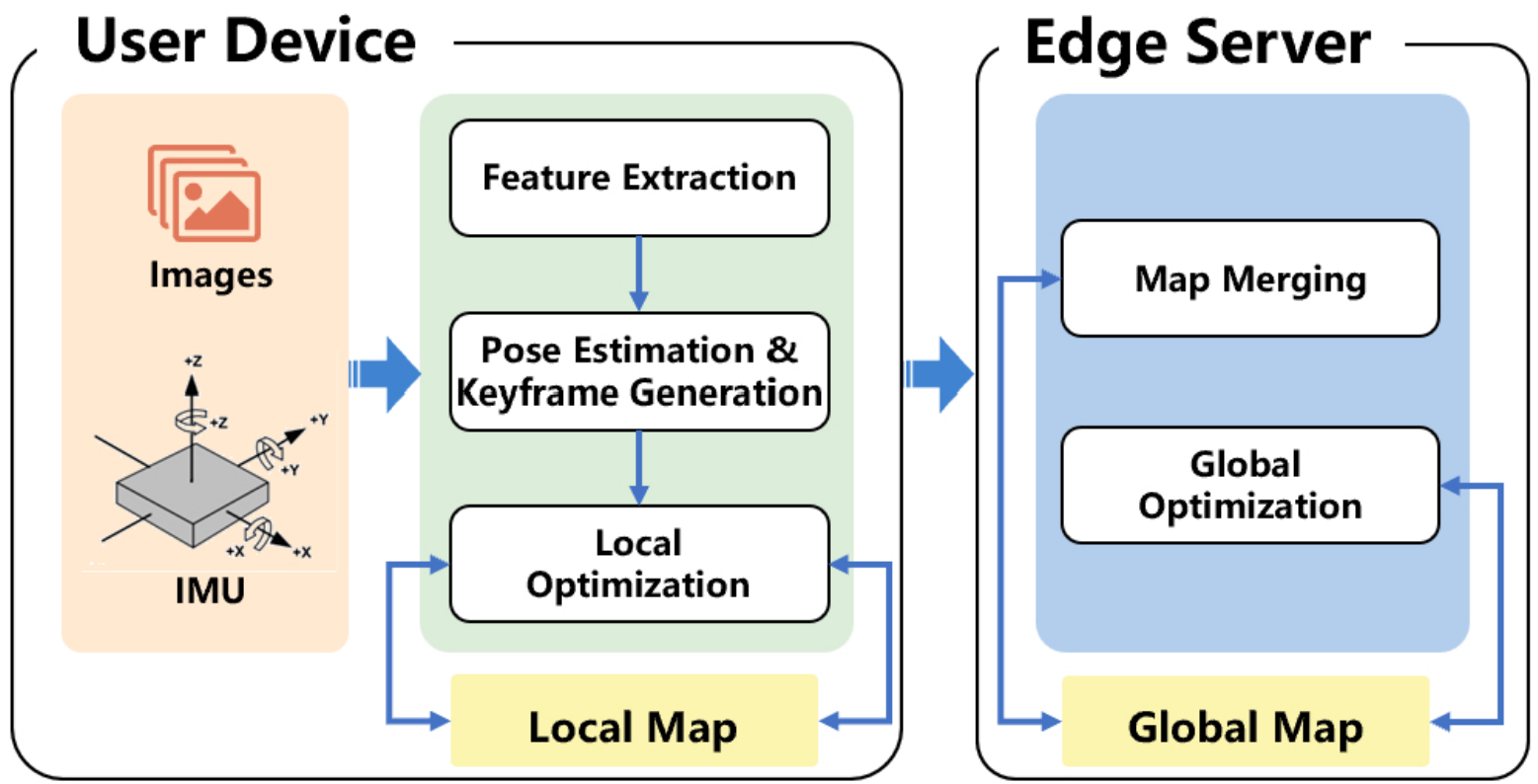} \vspace{-8pt}
\caption{\small\label{fig:orbslam}Overview of a vanilla shared-map architecture as discussed in ~\cite{covins}. Each user uploads raw data (in the form of keyframes) to the server. The server merges the map from different users and conducts global optimization. }
\vspace{-15pt}
\end{figure}

\begin{figure*}[!t]
\centering
\includegraphics[width=0.84\linewidth] {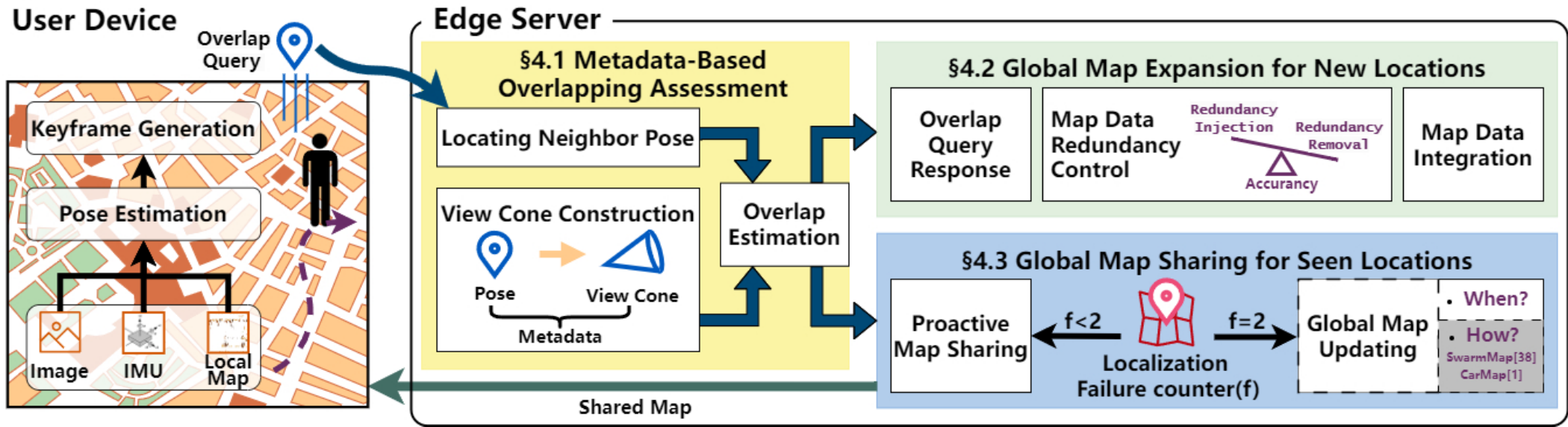}
\vspace{-11pt}
\caption{\small\label{fig:overview}\systemname Overview. The gray part (how to update) is not part of this work, and citations are given for further reference.}
\vspace{-10pt}
\end{figure*}

\vspace{-4pt}
\section{Visual SLAM Primer} 
\label{sec:prelim} \vspace{-2pt}
\noindent\textbf{ORB-SLAM3 Algorithm.} We use ORB-SLAM3~\cite{orbslam3}, the state-of-the-art visual SLAM algorithm, to explain how visual SLAM works. 
As shown in Fig.~\ref{fig:orbslam}, a distributed ORB-SLAM3 system consists of the following modules, namely, feature extraction, pose estimation, keyframe generation, local optimization, global optimization, and map merging. The first three modules are commonly referred to as tracking.

 \noindent\textbf{$\bullet$ Feature Extraction and Pose Estimation.} 
ORB-SLAM3 estimates the pose from 2D images as opposed to relying on GPS signals.
When the device's camera captures a frame, the SLAM algorithm first extracts its 2D ORB features~\cite{rublee2011orb} such as corners to distinguish unique characteristics in this frame.
These features are matched with those previously extracted, enabling the pose estimation algorithm to estimate the distance traveled since the previous frame, providing an initial assessment of the camera's current position and orientation, known as its \textit{\textbf{pose}}.  
The SLAM algorithm improves the pose estimation through an optimization procedure. Once the pose is determined, the algorithm projects 2D features into the 3D space, generating \textit{\textbf{map points}} for that frame.

 \noindent\textbf{$\bullet$ Keyframe Generation.}
To alleviate the mapping overhead, the frames that lack distinguishable features will be excluded from the follow-up mapping tasks.  In situations where the current frame's features do not closely align with those of the preceding frame (such as when the matching coefficient falls below a predetermined threshold), the algorithm examines the distance between these consecutive frames. If this distance is substantial, the current frame is designated as the \textit{\textbf{keyframe}}.
Mathematically, a keyframe $\textbf{K}$ consists of $n_f$ ORB features and $np$ 
map points: $\textbf{K} =  ( {\mathbf{P}}, F_0,F_1, \cdots,F_{n_f}, MP_0, MP_1, \cdots, MP_{np})$, 
where $\mathbf{P}$ denotes the pose matrix, $F_i$ denotes the $i$-{th} ORB feature,  and  $MP_j$ denotes the $j$-{th} map point. Following ORB-SLAM3, we set $n_f$ to 1000 for all images, irrespective of their resolutions, and determine $np$  dynamically for each keyframe based on how much the new keyframe overlaps with previous ones.

 \noindent\textbf{$\bullet$ Local Optimization.} Each selected keyframe with associated map point data is combined into the \emph{\textbf{map}}. The SLAM algorithm then applies local Bundle Adjustment (BA)~\cite{ba} to optimize neighbor keyframes within the map, refining their poses and map point estimation based on spatial constraints between them (such as observations among keyframes, mapping relationships from 3D map points to 2D features, etc). This local-area optimization is known as local optimization.

 \noindent\textbf{$\bullet$ Global Optimization.}
As more keyframes are inserted into the map, keyframe duplication is likely to happen, which forms a loop.  The SLAM algorithm then applies a global BA alignment to refine the poses of all the keyframes within the loop and associated map points, which can substantially reduce the accumulated errors.

 \noindent\textbf{$\bullet$ Map Merging.} Moreover,
if keyframes in one map resemble keyframes from another map, the two maps will be merged within a global coordinate system. The process involves detecting similarities between keyframes and calculating their relative poses. 

\vspace{2pt}\noindent\textbf{User-Participatory Shared Map Architecture}. Our system adopts a shared map architecture, as shown in Fig.~\ref{fig:orbslam}, where each participating device maintains a \textit{\textbf{local map}} to support pose estimation and keyframe generation. With this local map intended for devices, the user device performs tasks such as localization (calculating both position and orientation, totaling 6 degrees of freedom (DoF)) and autonomous navigation (no human intervention needed).
The map data created by each participating device will be transferred to a central edge node or a cloud server. To support long-term operation, the server maintains a \textit{\textbf{global map}} that combines maps from all participating devices through map merging. 

There are two main bottlenecks of such a user-participatory mapping system. The first bottleneck is the resource bottleneck on the server, especially the CPU computation resources and long-term memory cost. The second bottleneck stems from the resource consumption on the mobile device, including mobile data, CPU, battery power, etc., which limits the willingness or the extent of user participation. For instance, when serving 20 users (total traveling a 2604m trajectory), the vanilla shared-map system that has each user directly upload their keyframes requires 3.98GB of memory and takes 76 minutes to optimize the global map on an AMAX server equipped with two AMD EPYC 7H12 CPUs.  
It consumes 1GB of memory and 2.46W of power, and generates 114MB of network traffic for a user to map a 260-meter path on NVIDIA AGX Xavier. 
More importantly, in the real world, most users travel on trajectories that are already mapped, but there is no mechanism for them to utilize the existing map. Instead, they still spend a large amount of resources on building the map from scratch.  
\vspace{-4pt}\section{Overview of \systemname} \label{sec:overview} \vspace{-2pt}

\subsection{Design Goals}
\label{ss:design_goals} \vspace{-2pt}
\systemname is designed to support efficient and user-friendly visual SLAM with the following goals.

\noindent\textbf{High scalability}. 
\systemname should be able to support large numbers of participants under given networking, computing, and memory constraints. 

\noindent\textbf{Minimal redundancy}. \systemname should eliminate redundant map data transmission while ensuring map accuracy and coverage. This allows for the effective utilization of limited computing, networking, memory, and mobile resources.

\noindent\textbf{Continuous updates}. To keep the map up to date, \systemname should support updating over an extended period of time.

\vspace{-6pt}
\subsection{System Overview} \vspace{-2pt}
\label{ss:system_overview}
Following the old wisdom that ``it takes a village to raise a child'', \systemname offloads the task of map generation to map users, progressively constructing the global map when and where the service is needed.
The overall system consists of three design components, as shown in Fig.~\ref{fig:overview}. 

\noindent\textbf{Metadata-Based Overlapping Assessment}. 
Upon generating a new keyframe, the mobile device synchronizes with the server by querying the degree of overlap with the global map without having to upload the bulky keyframe data to the server, which is approximately 160KB for all image resolutions. This lightweight synchronization reduces the device's mobile data and battery consumption as well as conserves the server's computing/memory resources. 

\noindent\textbf{Global Map Expansion for New Locations}.   
If the degree of overlap is lower than a predetermined threshold, the location is identified as new or partially new. 
In this case, the server notifies the device whether the entire keyframe needs to be uploaded or only a portion of it.  
Using the uploaded ``fresh'' data, the server then expands the global map suitably. 

\noindent\textbf{Global Map Sharing for Seen Locations}. When the degree of overlap surpasses the threshold, \systemname recognizes the location as pre-existing and no longer requires mapping. Instead, \systemname distributes the map surrounding the location to the device. Subsequently, the device can replace its local map with the one obtained from the server to significantly conserve resources on mobile devices. 
\vspace{-4pt}\section{System Design} \label{sec:design}

\vspace{-2pt}
On the mobile device, as each frame is captured, \systemname conducts feature extraction, pose estimation, and keyframe generation. 
Hereafter, a metadata-based global map overlapping assessment algorithm is engaged to determine the extent of overlap between the new keyframe and the global map (\S\ref{sec:assess}). Depending upon the overlap situation,  the server either requests fresh map data to expand the global map (\S\ref{sec:new}) or shares the suitable map segment with the user (\S\ref{sec:seen}),
illustrated in Fig.~\ref{fig:overview}.

\vspace{-4pt}
\subsection{Metadata-Based Overlap Assessment}\label{sec:assess} 
\begin{figure}[!t]
\centering
\includegraphics[width=.85\linewidth] {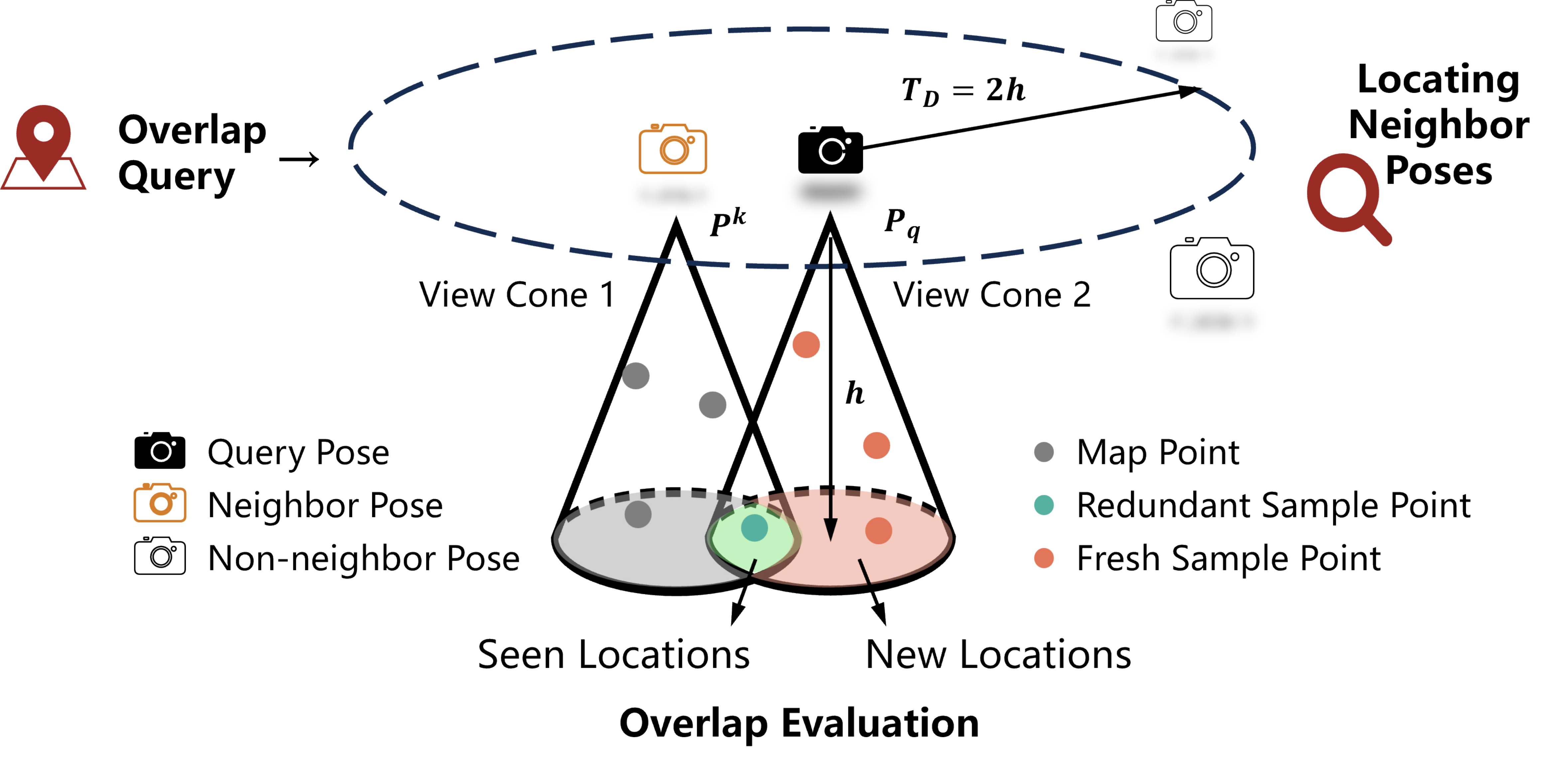}  \vspace{-12pt}
\caption{\small\label{fig:frustum}Illustration of the camera pose, its view cone, and the overlap evaluation between two cones. The sampling points marked as FRESH or REDUNDANT are also included.}
\vspace{-10pt}
\end{figure}

\vspace{-2pt}
When a new keyframe is generated, it is necessary to synchronize with the server to evaluate the overlap between the keyframe and the global map. To optimize resource usage during this synchronization process, we propose utilizing two metadata attributes of a keyframe: its spatial index (represented by its pose) and its spatial range (represented by its 3D view cone). Fig.~\ref{fig:frustum} illustrates this idea.
Below we elaborate on the metadata-based overlap assessment process. 

\noindent\textbf{Overlap Query and Query Pose.} First, the device sends an overlap query to the server for overlap assessment. The query is defined as: \vspace{-2pt}
\begin{equation} \small \vspace{-2pt}
Overlap\_Query = \{C, K, \mathbf{P_q}\}, 
\end{equation}
where $C$ and $K$ represent the user ID and keyframe ID, respectively, and 
$\mathbf{P_q}$ is the keyframe's pose, referred to as the \emph{query pose}. We also refer to the keyframe as the query keyframe for the sake of convenience.  
The query packet size is 64 bytes, which is three orders of magnitude smaller than the keyframe packets (about 160 KB for all image resolutions). As the number of keyframes increases, a significant amount of mobile data and battery power is saved. 

\noindent\textbf{Constructing the 3D View Cone for a Given Pose.} 
Once the query packet arrives at the server, \systemname takes the query pose $\mathbf{P_q}$ and constructs the corresponding 3D view cone that represents the extent of the viewing range from the pose.  The field of the 3D view cone is depicted in Fig.~\ref{fig:frustum}.
Given a pose $\mathbf{P} = (x, y, z, \theta_{roll}, \theta_{pitch}, \theta_{yaw})$, where $(x, y, z)$ denotes the camera position; $(\theta_{roll}, \theta_{pitch}, \theta_{yaw})$ denotes the camera rotation in the global coordinate (the global coordinate is discussed in Sec.~\ref{sec:new}). 
Its view cone is calculated as follows. Firstly, a 3D optical coordinate centered at $(x, y, z)$ is established with the rotation of the 3D view cone as $(\theta_{roll}, \theta_{pitch}, \theta_{yaw})$.
Then, the camera's field of view parameters  ${FOV}$ are utilized to determine this 3D view cone, which is defined as below: \vspace{-2pt}
\begin{equation}\small \vspace{-4pt}
{FOV}  = \max\{2\arctan(\frac{c_x}{f_x}) ,2\arctan(\frac{c_y}{f_y})\},
\label{eq:fov} 
\end{equation}
where $f_x$, $f_y$, $c_x$, and  $c_y$   are the intrinsic parameters of the camera associated with the pose. Specifically, $f_x$ and $f_y$ are the camera's focal lengths in the horizontal and vertical directions, respectively, and  $c_x$ and $c_y$ represent the horizontal and vertical coordinates of the optical center on the camera. 

Finally, combined with a height, $h$, the 3D view cone ${VCone}_{P}$ is determined. Here, we set $h = 20$ meters following the convention in ORB-SLAM3~\cite{orbslam3}: 
\vspace{-2pt}
\begin{equation}\small \vspace{-3pt}
{VCone_p}  = \{\mathbf{P}, h, {FOV}\}.
\end{equation}
Each 3D view cone represents the corresponding field of view based on the camera's intrinsic parameters.  Once the view cone ${VCone}_{P}$  is generated for $\mathbf{P}$,  all map points that can be seen from the pose are included in the cone. Fig.~\ref{fig:frustum} shows a camera pose and its corresponding view.

\noindent\textbf{Locating Neighbor Poses in the Global Map.}   
Once the query packet arrives at the server, \systemname takes the query pose $\mathbf{P_q}$ as the reference to find neighbor map frames in the global map that `see' the same map point with $\mathbf{P_q}$. 
This is achieved by comparing the angle and Euclidean distance between their poses. 
Before discussing the details, we first define the global map.  The global map $\mathbf{M}$ consists of all the map frames with each frame indexed by its pose: \vspace{-2pt}
\begin{equation} \footnotesize \vspace{-2pt}
\mathbf{M} =
    \begin{cases}
  (  \mathbf{P^0},  F^0_0, F^0_1,   \cdots, F^0_{n_f},  &MP^0_0, MP^0_1, \cdots, MP^0_{{np}_{0}} ),   \\
 (  \mathbf{P^1},  F^1_0, F^1_1,   \cdots, F^1_{n_f},  &MP^1_0, MP^1_1,   \cdots, MP^1_{{np}_{1}} ),   \\
   &\cdots   \\ 
 ( \mathbf{P^k}, F^k_0, F^k_1,   \cdots, F^k_{n_f},   &MP^k_0, MP^k_1,   \cdots, MP^k_{{np}_{k}} ), 
    \end{cases}
 \end{equation}
\noindent where each entry in the database denotes a map frame $k$, characterized by the pose $\mathbf{P^k}$, 2D features $F^k_{0}, \cdots, F^k_{n_f}$, and associated map points $ MP^k_{0}, \cdots, MP^k_{np_{k}}$.  

To identify neighbor poses, we first select poses that are located within a certain distance range from $\mathbf{P^k}$. From these poses, we further select those that have a similar orientation angle to $\mathbf{P^k}$. We define this process as below: \vspace{-2pt}
\begin{equation} \small \vspace{-2pt}
\begin{aligned}
\mathbf{S_{n1}} = \mathbf{P^k}\in \mathbf{M} \mid   dist (\mathbf{P_q} , \mathbf{P^k}  )  &< T_D,
\end{aligned}  \vspace{-4pt}
\end{equation} 
\begin{equation} \small \vspace{-3pt}
\begin{aligned}
\mathbf{S_{n2}} = \mathbf{P^k}\in \mathbf{S_{n1}} \mid  angle ( \mathbf{P_q}  ,  \mathbf{P^k} ) &< \frac{1}{2}({FOV_q}+{FOV^k}), 
\end{aligned}
\end{equation}
where $\mathbf{S_{n1}}$ and $\mathbf{S_{n2}}$ are the selected pose sets, $T_D$ is the distance threshold (we have $T_D = 2h$ as illustrated in Fig.~\ref{fig:frustum}), and $FOV_q$ and $FOV^k$ are fields of view of the query pose and map frame pose, respectively.
Subsequently, we obtain the corresponding neighbor frames as well as the neighbor points (i.e., points that are contained in the neighbor frames). To evaluate the overlap between the query keyframe and the global map, we need to further examine the spatial relationships between neighbor points and the query keyframe. 

\begin{figure}[!t]
\centering
\includegraphics[width=.83\linewidth] {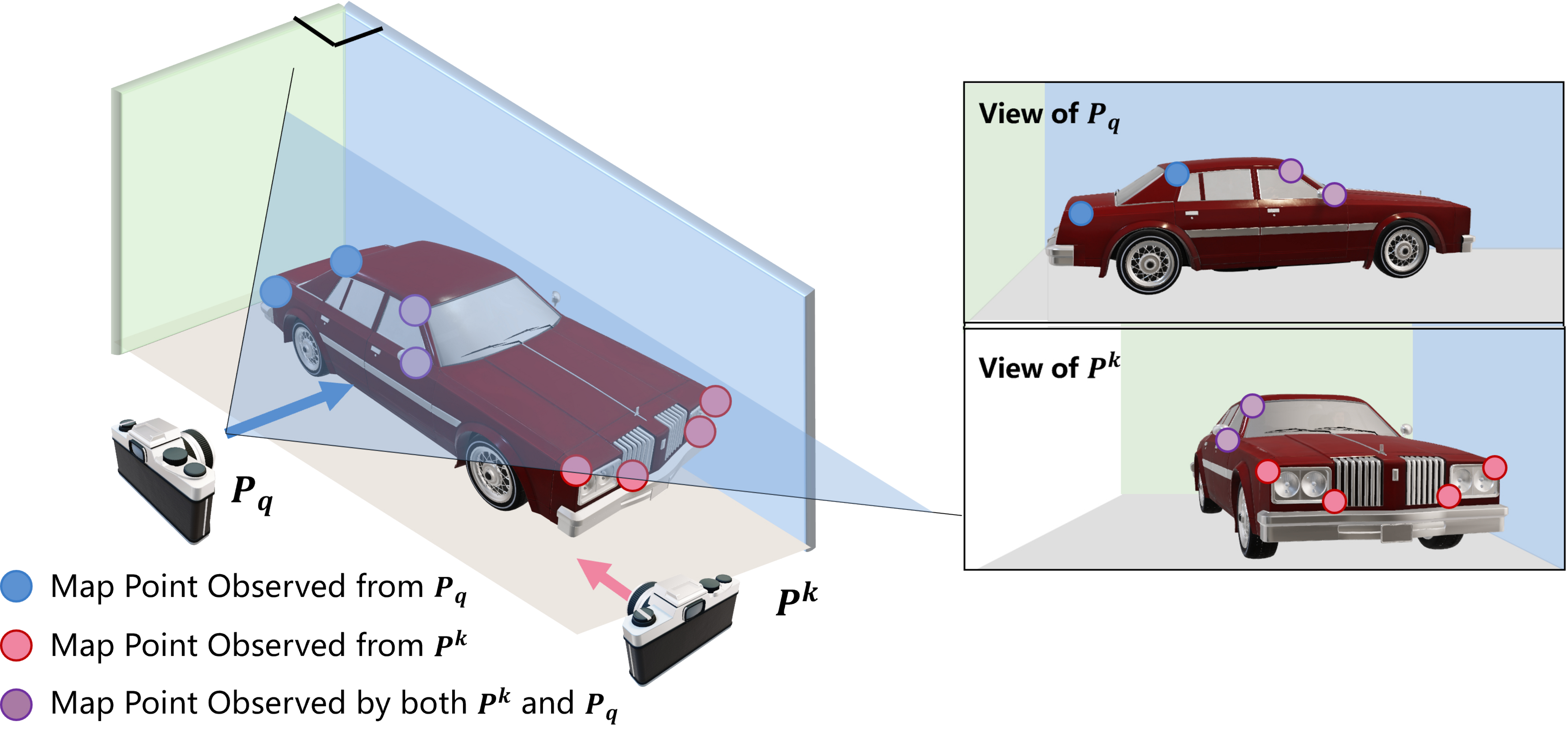}
\vspace{-10pt}
\caption{\small\label{fig:view}This example illustrates that neighbor map points may fall within the query pose's view cone, but should not be deemed as REDUNDANT. Thus, we cannot simply detect the overlap by evaluating how many neighbor points on the map fall within the view cone of the query pose. }
\vspace{-12pt}
\end{figure}

\noindent\textbf{Overlap Estimation through Hierarchical Searching.}  
Overlap evaluation in the 3D space is a tricky issue. Indeed, in 3D environments, single-view observations are insufficient to reconstruct the scene, necessitating contributions from multiple views by different users. This point is illustrated in Fig.~\ref{fig:view}. As such, we cannot identify the overlap area between the query keyframe and the global map by simply examining whether the neighbor map points fall within the 3D view cone ${VCone}_{P}$   -- even if a point falls within the 3D view cone, it might be captured from a different view on a different object surface and is not redundant with the query keyframe. Therefore, such a view-cone-level overlap estimation likely leads to overestimation.   
  
To address this challenge, we devise a fine-grained map-point-level estimation method. Specifically, We generate $K$ sampling points ($K$ is the number of map points within a keyframe) uniformly spaced within ${VCone}_{P}$, with $r$ denoting the mean distance between two adjacent sampling points. 
Given a sample point, if we can find map points located within a radius of $r$, we consider this sampling point REDUNDANT. A naive linear traversal scheme has a complexity of $O(K*m)$ with $m$ map points, which is quite CPU-intensive.    
In \systemname, we leverage a KD-tree-based 3D map point indexing structure, which facilitates hierarchical search space partitioning for efficient indexing. We build a neighborhood KD-tree that includes all the neighbor map points. For each sample point, the KD-tree search complexity is $O(log_2m)$ with $m$ map points.

\systemname uses the ratio of REDUNDANT points over the entire sampling points as the proxy for the overlap degree of the query keyframe with respect to the global map.  If the overlap degree exceeds a predefined threshold $T_{seen}$, the query pose is marked as a ``seen'' location. 
In our implementation, we set $T_{seen}$ as 90\%, following the setting of ORB-SLAM3~\cite{orbslam3}, which uses this threshold for keyframe culling --  it deems a keyframe as redundant if more than 90\% of its map points are observed by other keyframes.

\vspace{-4pt}
\subsection{Map Expansion for New Locations} \label{sec:new} \vspace{-2pt}
When the query keyframe does not overlap or has minimal overlap with the global map, \systemname considers the query pose to be a new or partially new position. In such a scenario, \systemname requests the device to send its map data to the server for map expansion. 

\noindent\textbf{Overlap Query Response.} The server responds to the overlap query by returning the detailed sampling results. Specifically, after sampling the query pose's view cone $VCone_P$, each sampling point is identified as REDUNDANT or FRESH, represented by symbols $PT_{redundant}$ and $PT_{fresh}$, respectively. If there are fewer $PT_{redundant}$ than $PT_{fresh}$ (meaning the query keyframe has a less degree of overlapping with the global map), the server returns the list of $PT_{redundant}$, with an additional status bit indicating they are $PT_{redundant}$. Otherwise, the server sends the list of $PT_{fresh}$ and marks the status bit accordingly. The response message is defined as:  \vspace{-2pt}
\begin{equation}\small \vspace{-2pt}
\small Overlap\_Response = \{ S,  PT_0, PT_1, \cdots,  PT_{n_S}  \}, 
\end{equation}
where $S$ is the status bit,  $PT_0, \cdots,  PT_{n_S}$ are the corresponding sampling points.

\noindent\textbf{Map Data Redundancy Control.} 
Once the list of sampling points is received,  the device prepares the map data to be sent to the server. Below, we explain this process assuming sampling points to be $PT_{redundant}$.  
For each $PT_{redundant}$, the device retrieves the map points that are within a radius of $r$ (the mean distance between two sampling points) and deletes them from the keyframe, leaving only fresh map points for the global map. The device subsequently sends them along with the corresponding descriptors to the server. 
This mechanism, referred to as \emph{redundancy removal},  ensures that the server only receives non-redundant data. 

Though the above minimal-redundancy policy is computation/transmission efficient, it can be harmful to the map quality as the global optimization function often relies on having some redundant data from which additional optimization constraints can be derived~\cite{redundancy}. 
An imprecise global optimization yields incorrect poses on the server, leading to overlap assessment errors.
Therefore, we also devise a simple yet effective \emph{redundancy injection} mechanism to introduce a small amount of redundant data suitable for global optimization. 
Specifically, we record how many times a map point is observed by the device and transmit those points that are observed more often than average to the server. By uploading these map points, we can improve the map quality because they can provide more dependable constraints for optimization among keyframes. 
Meanwhile, the system cost is still kept close to the minimum.

\noindent\textbf{Map Data Integration and Coordinate Alignment.} \systemname supports asynchronous participation by integrating the map data chronologically. When the new map data are received by the server, they are integrated into the global map.  
The server first organizes the map data into a new map frame and then inserts this frame according to the pose:  \vspace{-2pt}
\begin{equation} \footnotesize  \vspace{-2pt}
 \begin{aligned} 
  Map\_Insert( \mathbf{M}, \mathbf{P^i},  F^i_0, F^i_1,  & \cdots, F^i_{n_{new}}, MP^i_0, MP^i_1, \cdots, MP^i_{np_{new}} ),   \\ \vspace{-1pt}
  n_{new} &\leq n_f , np_{new} \leq  np_{i},
  \end{aligned} 
\end{equation}
where $\mathbf{P^i}$ denotes the pose to be inserted, $F^i_0, \cdots, F^i_{n_{new}}$ denotes the ORB features, and $MP^i_0,  \cdots, MP^i_{np_{new}}$ denotes associated map points. 
Even though the number of ORB features and the number of map points are likely smaller than those in a regular keyframe due to overlap removal, we still allocate the same memory each time to facilitate easy map updates. 

When a user joins \systemname, the system continuously monitors the alignment between the new user's map and the global map. Upon successful alignment, a map merging process is triggered, wherein the new user's map on the server is combined with the global map. We need to align the coordinate systems from different cameras into a unified coordinate system. We follow the approach in ORB-SLAM3~\cite{orbslam3} and Covins~\cite{covins} to align the coordinate systems, mainly involving estimating the rotation and translation transformation parameters. In \systemname, we initiate a global optimization procedure when a user concludes their session.

\begin{figure}[!t]
\centering
\includegraphics[width=.87\linewidth] {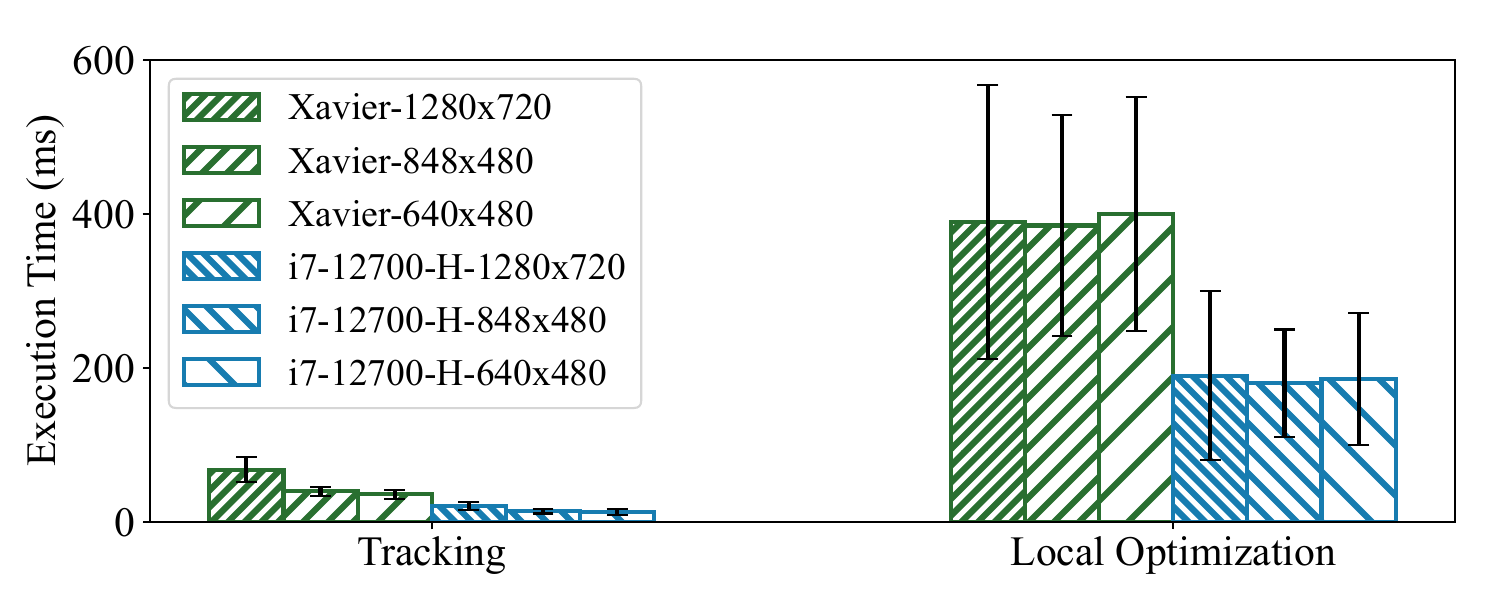} 
\vspace{-10pt}
\caption{\small\label{fig:local} The execution times of tracking and local optimization of different image resolutions. The experiments are conducted on an NVIDIA Jetson AGX Xavier with an 8-core Carmel CPU running at 2.2GHz, and a Lenovo  Y9000p laptop with a 14-core i7 12700-H CPU running at 2.4GHz. }  
\vspace{-12pt}
\end{figure}

\vspace{-4pt}
\subsection{Map Sharing for Seen Locations} \label{sec:seen} \vspace{-2pt}
When the overlap degree between the query keyframe and the global map exceeds the threshold $T_{seen}$, \systemname obtains the shared map by retrieving the nearby map frames, groups them into packets, and dispatches them to the query device. The device then employs the shared map data as its local map to avoid performing local optimization operations. Fig.~\ref{fig:local} shows the measured latency for tracking and local optimization functions on different platforms with varying image resolutions. It is observed that the local optimization procedures require 400 milliseconds on average even with images of resolution $640 \times 480$ with the NVIDIA Xavier platform. 
By eliminating these operations, \systemname significantly reduces the computation overhead on mobile devices.

\noindent\textbf{Proactive Map Sharing.} An intuitive approach is to share all the map points that fall within the 3D view cone $VCone_P$ with the query device. However, with this conservative map-sharing approach,  the device needs to repeatedly conduct the overlap query for each subsequent keyframe, which is quite costly. 
In \systemname, we instead propose a proactive sharing approach that allows the server to ``look ahead'' -- sharing extra map data to the device. By doing so, \systemname can minimize the queries from the device and thus cut down the overlap assessment effort on the server. 
Here, we represent the shared map with a 3D view cone  $Shared\_VCone_{{P}}$: 
\begin{equation} \small \vspace{-2pt}
 Shared\_VCone_{P}   =  \{\mathbf{P_q},  h_{q},  \alpha * FOV_q \},
\end{equation}
where $\alpha$ denotes the oversharing factor. In the evaluation, we demonstrate how a slightly larger oversharing factor leads to a lower overhead on the user device.

\noindent\textbf{Localization Failure with the Shared Map.} Once the shared map is received, the user replaces the local map with the shared map and modifies its subsequent behavior as follows. When a new camera frame is captured, it performs regular pose estimation tasks with the shared map. In this process, when the number of matching points between the new frame and the shared map drops below a certain point (75 in ORB-SLAM3), it is deemed as a localization failure, and a new keyframe is generated. At this point, the device resumes its overlap query and stops using the shared map.

\noindent\textbf{Detecting the Need for a Global Map Update.} 
When the user resumes the overlap query due to localization failure, it needs to discriminate the following two cases: (1) the user has moved out of the shared map, or (2) the actual environment has changed. In \systemname, we adopt the following simple check mechanism. As shown in Algorithm~\ref{alg:code}, \systemname firstly requests a shared map from the server, if the received map is not empty, the user will try to localize on the shared map.  If  $f$ consecutive failures have been experienced (we pick $f=2$ in our implementation for more timely updates), the user will send actual keyframe data along with its overlap query. Once receiving such a query, the server compares the query keyframe with the corresponding map data to examine whether an update operation is required. For example, we can leverage the users' trajectory to estimate map confidence.  For each high-confidence map point and its $K$ nearest neighbors, the server assesses their geometric similarity in relation to the newly received map data and evaluates the need for a map update.
We will show its effectiveness in Sec.~\ref{sec:results}.
To more precisely eliminate the impact of temporal obstacles, a viable solution is to add an object recognition module to determine whether to send the current frames to the server or not.
If the server concludes that the environment has changed, the previously proposed updating mechanism such as SwarmMap~\cite{xu2022swarmmap} can be adopted for efficient map updating. 

          \begin{algorithm}[t] \small 
		  \caption{\label{alg:code}Device-Initiated Global Map Update Detection}  
		\begin{algorithmic}[1]
			\Statex \textbf{Input:} User ID C, Keyframe ID K, Query pose $P_q$,  Local keyframe list $KFs$, Maximum iteration number $f$.   
                    \For{ $i=0$ to $f$} 
                        \State  $M = Request\_Shared\_Map(C,K,P_q)$   \Comment{ \textcolor{codecolor}{\S\ref{sec:seen}: Map Sharing} }
                        \If  {  $M \neq NULL$ }         \Comment{  \textcolor{codecolor}{\S\ref{sec:seen}: Localization } }
                             \If  { $Localize(M,P_q) == SUCCESS $ } 
                                 \State   return;
                            \EndIf
                        \Else  \Comment{  Case 1 }
                             \State  $Map\_Expansion(C,K,P_q,M)$   \Comment{  \textcolor{codecolor}{\S\ref{sec:new}} }
                             \State  return;
                        \EndIf
                    \EndFor

                 \State  S = $Get\_Update\_Status(KFs)$  \Comment{ \textcolor{codecolor}{ \S\ref{sec:seen}: Map Updating} }
                   \If  { $ S == EXPANSION $ }   \Comment{  Case 1 }
                                 \State   $Map\_Expansion(C,K,P_q,M)$
                  \ElsIf{$ S == UPDATING $}     \Comment{  Case 2 }
                                \State    $Map\_Update(KFs)$  
                  \EndIf
		\end{algorithmic} 
	   \end{algorithm} 
    
\vspace{-4pt}
\section{Implementation} \vspace{-2pt}
 
We have implemented a prototype \systemname system in C++.
\systemname contains both the client end and server end, solely CPU-based. The client end is developed based on ORB-SLAM3~\cite{orbslam3}, while the server end is based on Covins~\cite{covins}. 
Different from Covins, which only supports a vanilla shared-map system, \systemname strives to minimize the system overhead of such a system for both the server and the user devices by minimizing data redundancy when expanding the map, and facilitating map sharing for users who travel on similar paths.
\systemname added approximately 3,100 lines of  C++ code on top of Covins. 
We adopt ZeroMQ~\cite{hintjens2013zeromq}, an asynchronous messaging library widely used in distributed systems to develop the communication module in our system. 
\systemname made no modifications to core SLAM functions, hence it can transform any point-based SLAM system into a user-participatory SLAM system. Below we list five key functions in \systemname.
 
\vspace{4mm}
\noindent\fbox{%
\small
\parbox{0.45\textwidth}{%
 \textcolor{codecolor}{SendMetaData}(OverlapQuery) $\rightarrow$ void;\\ 
 \textcolor{codecolor}{AssessOverlap}(Map, OverlapQuery) $\rightarrow$ QueryResponse;\\
 \textcolor{codecolor}{PartitionKF}(QueryResponse, KF\_orig) $\rightarrow$ KF\_new;\\ 
 \textcolor{codecolor}{MapInsert}(Map, KF\_new) $\rightarrow$ Map;\\
 \textcolor{codecolor}{RequestSharedMap}(OverlapQuery) $\rightarrow$ Map.
}}\vspace{4mm}

We opted to implement the system prototype on NVIDIA AGX Xavier instead of smartphones to expedite the research process since ORB-SLAM3 depends on many third-party libraries that are native to Linux. 
Each Xavier is equipped with an 8-core CPU Carmel running at 2.2GHz and has similar CPU performance to the Arm’s Cortex-A75 inside the Snapdragon 845 in Google Pixel 3~\cite{ANANDTECH}. 
Given the current memory and computing overhead of \systemname, we believe our client-side programs can work on COTS smartphones as well.
For the map server, we use an AMAX server, equipped with two AMD EPYC 7H12 CPUs running at 2.6GHz with 1 TB DDR4 RAM.
The device communicates with the server through Wi-Fi links on 5GHz frequency band. 
The measured upstream and downstream bandwidth are 21.3MB/s and 23.8MB/s.

\begin{figure*}[!t]
\centering
\begin{tabular}{cc}
\includegraphics[width=.46\linewidth] {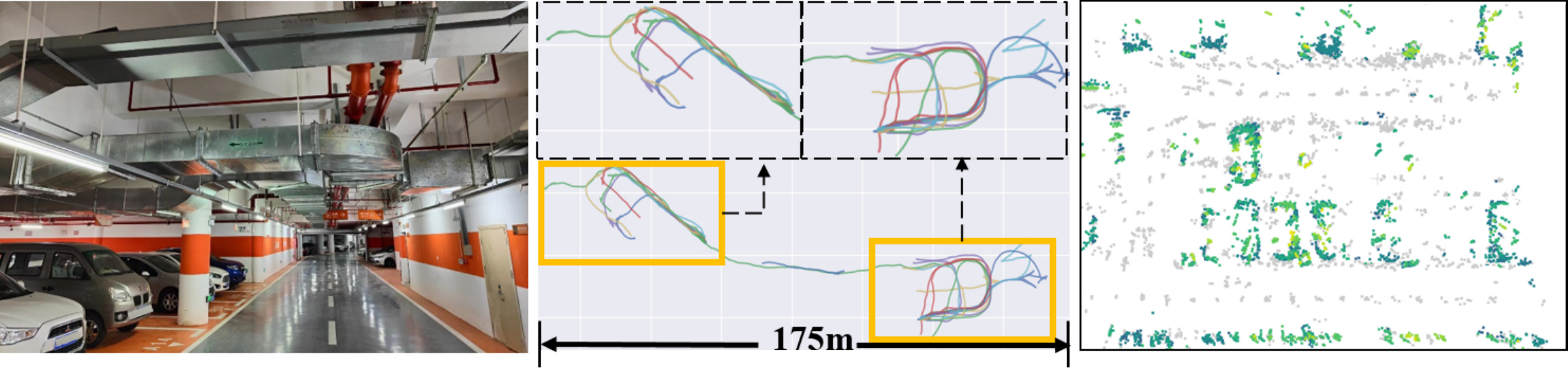}\hspace{-0.2cm} & 
\includegraphics[width=.46\linewidth] {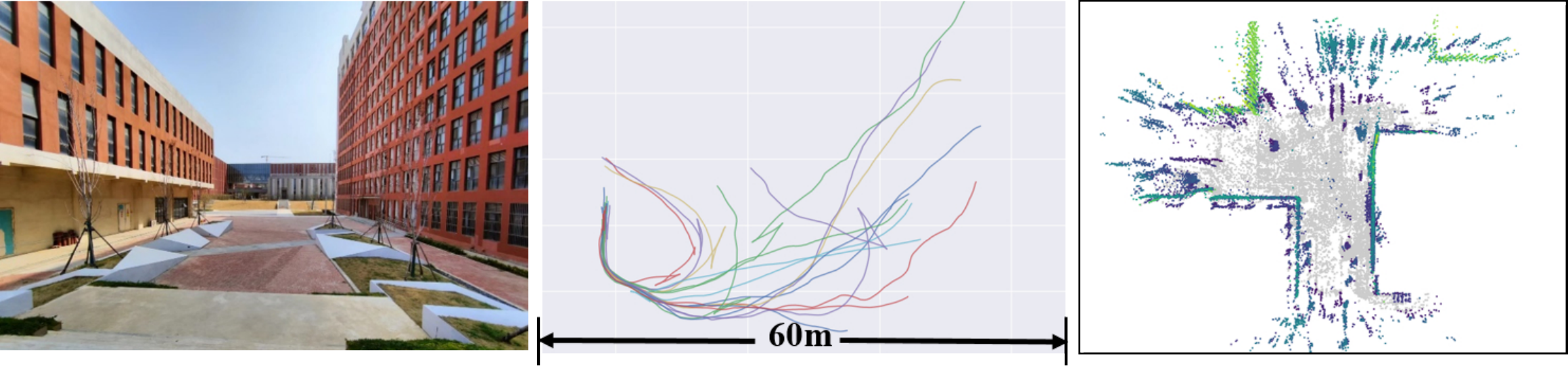} \vspace{-5pt}\\
\small(a) \datasetgarage & \small(b) \datasetbuilding\\
\includegraphics[width=.46\linewidth] {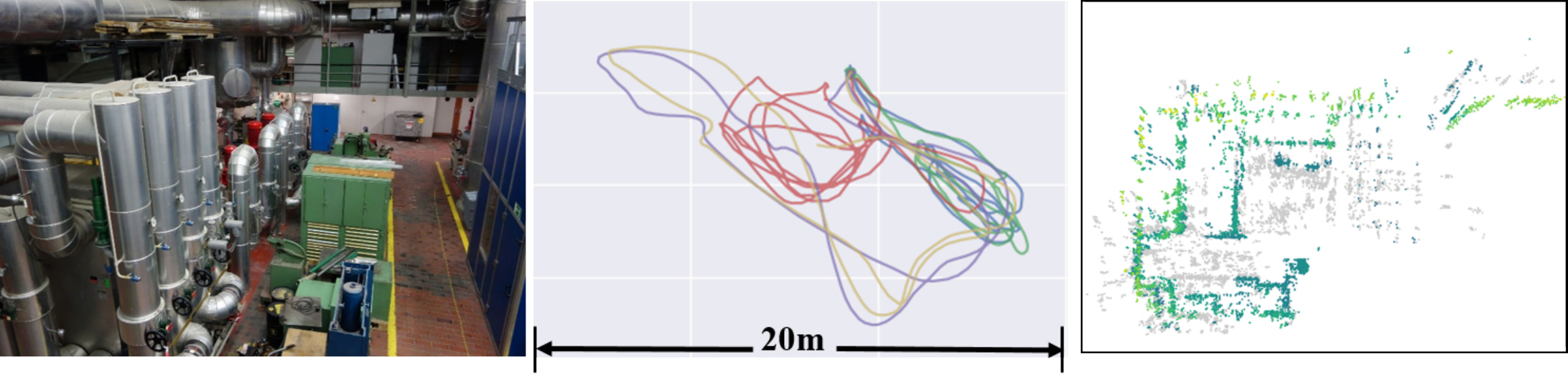} \hspace{-0.2cm} &
\includegraphics[width=.46\linewidth] {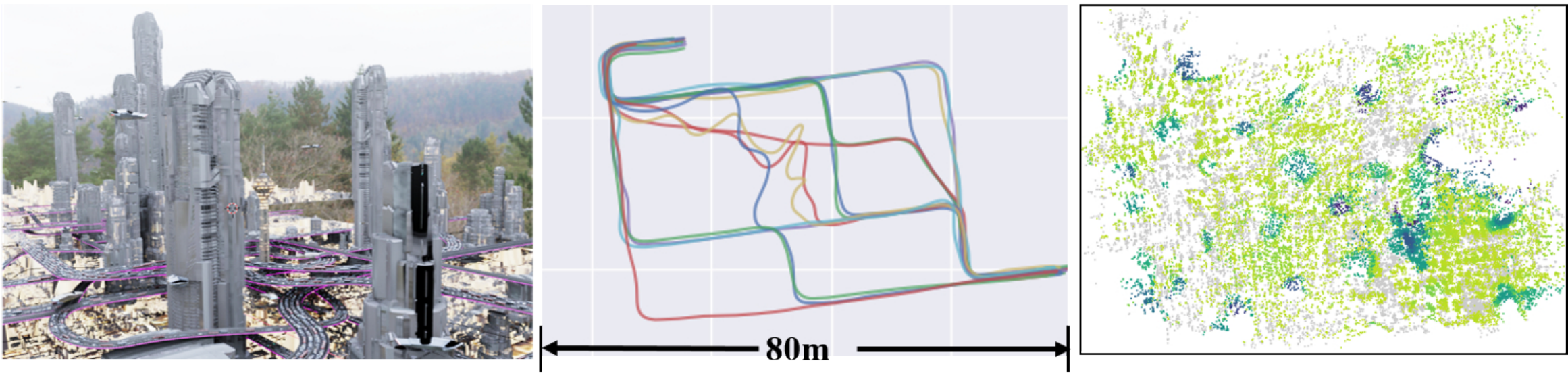} \vspace{-5pt}\\
\small(c) EuRoC & \small(d) Future-City\\ 
\end{tabular}
\vspace{-12pt}
\caption{\small\label{fig:traj} The four evaluation settings, the data-collection trajectories and the reconstructed 3D maps are shown here. The total length of the trajectories for each scenario are as follows: 2604m for Indoor-Garage, 2686m for Outdoor-Plaza, 453m for EuRoC, and 2358m for Future-City.}
\end{figure*}

\begin{table}[b]
\setlength\tabcolsep{4pt} 
 \renewcommand\arraystretch{1}
	\scriptsize \vspace{-8pt}
    \begin{center}
	\begin{tabular}{ |c|c|c|c|c| }
                \hline 
                  Setting    &  Resolution    &   \thead{\scriptsize  Distortion \\ \scriptsize $k_1$,$k_2$}   & \thead{  \scriptsize Exposure  \\    \scriptsize Time ($\mu$s) }   &    \thead{ \scriptsize Intrinsic  Parameters \\ \scriptsize $f_x,f_y,c_x,c_y$ }        \\
				\hline  \hline 
    	   \multirow{3}*{     \thead{  \scriptsize Garage \& \\ \scriptsize Plaza  }   }
			        &  1280x720    &  [-0.046,0.032]    & 80 &   [634.2,634.8,631.8,359.5]          \\   
                    \cline{2-5}
    	           & 848x480      &    [-0.044,0.034]  & 120 & [423.7,423.0,419.6,239.7]   \\ 
                   \cline{2-5}
    	            & 640x480    &    [-0.046,0.038]    & 120 &  [383.4,383.7,316.5,239.6]    \\
                    \cline{2-5}
                    \hline
    	        EuRoC   & 752x480  &   [-0.283,0.074]    & automatic &  [458.7,457.3,367.2,248.4]    \\
             	\hline
    	  FutureCity   & 752x480    &    [-0,0]    & - &  [455.0,455.0,376.0,240.0]    \\ 
    				\hline
    			\end{tabular}
       \end{center}
  \caption{\small\label{tab:camera} Data collection camera parameter settings. }\vspace{-10pt}
\end{table}

\vspace{-4pt}
\section{Evaluation}
\label{sec:results} \vspace{-2pt}
We have conducted a thorough evaluation comparing our proposed \systemname and the vanilla user-participatory SLAM system Covins. Our objective is to show that (1) \systemname can significantly reduce both device-side and server-side resource consumption for building such a shared-map system; (2) \systemname can offer low-cost map sharing when users travel on similar paths while facilitating timely map updating -- note that we focus on detecting when to update while schemes such as SwarmMap~\cite{xu2022swarmmap} focus on how to perform the updating operation; and (3) \systemname can deliver a comparable map quality at a much lower cost. 

The system evaluation consists of two phases: (1) data collection and (2) mapping experiments. In the first phase, camera and IMU data are collected in various scenes and recorded in the form of rosbag. We run \systemname on this collected data in the second phase.
We \emph{deliberately} designed this two-phase experimentation approach such that we can focus on the design and evaluation of the proposed protocols and algorithms. 
Firstly, we use ROS~\cite{ROS}, a communication middleware, to record sensor topics as rosbags. 
By playing back the recorded data, we can utilize the same set of data collected in one pass to drive system design, evaluation, and improvement, especially when compared with the baseline. 

\vspace{-4pt}
\subsection{Dataset Collection}\label{subsec:Dataset} \vspace{-2pt}
The evaluation is based on four distinct datasets, which comprise two manually assembled collections, a publicly available SLAM dataset, and a simulation dataset. 
We employed three types of sensors: Intel RealSense D455 depth cameras for monocular imaging, MTI-300-2A8G4 Xsens IMUs, and Robosense RS-Helios 1615 LiDAR to collect data. We use a LiDAR-based SLAM algorithm, PFilter~\cite{duan2022pfilter}, which is robust against lighting conditions and dynamic objects, to generate the ground truth.
We employ Kalibr~\cite{kalibr} to calibrate camera and IMU. Besides, taking into account the different frequencies of the camera and IMU, \systemname matches each image frame to the closest IMU data point whose timestamp is less than or equal to that of the image. Then, it calculates the integral of the high-frequency IMU data between two adjacent frames ($f_i$ and $f_{i+1}$) and aligns the results with frame $f_{i+1}$ accordingly.
To account for hardware heterogeneity, we gathered the data under different camera parameter settings, such as resolutions, distortion, exposure time, and focal lengths.  We list the parameter settings in Tab.~\ref{tab:camera}.
A total of 35 volunteers were invited to generate SLAM trajectories, including 20 map expanding users and 15 map sharing users.

\noindent\textbf{Dataset I: \datasetgarage}. 
The first testing field is an indoor garage ($45m\times175m$), as shown in Fig.~\ref{fig:traj}(a). 
The environment frequently undergoes abrupt changes due to the presence of infrared lights and reflective tiles on the floor, which result in substantial reflections.

\noindent\textbf{Dataset II: \datasetbuilding}. 
The second testing field is an outdoor open space ($60m\times45m$), as shown in Fig.~\ref{fig:traj}(b). 
This location represents the most challenging scenario out of the four testing fields,  as a result of its spacious layout, and intense lighting conditions.

\noindent\textbf{Dataset III: EuRoC}. 
EuRoC Micro Aerial Vehicle (MAV) Dataset~\cite{euroc_paper} is one of the most commonly used datasets for evaluating visual SLAM systems. 
We show them in Fig.~\ref{fig:traj}(c). 
We use five sequences collected in the industrial machine hall, with overlapping segments among them. 
 
\noindent\textbf{Dataset VI: \dataset}. 
Furthermore, we constructed a dataset utilizing a simulated mini-city ($80m\times75m$) following Covins~\cite{covins}, as shown in Fig.~\ref{fig:traj}(d). The dataset comprises 20 distinct trajectories of a drone. 

\vspace{-4pt}
\subsection{Evaluation Metrics}\label{subsec:metrics} 
\vspace{-2pt}
In our evaluation, we mainly report the following metrics. 
\noindent\textbf{Resource Consumption on Devices}. We report the traffic volume in $KB$ per keyframe to evaluate mobile data consumption, including overlap query messages (upload traffic), map data response messages (download traffic), and map data upload messages (upload traffic). Additionally, we report the CPU utilization (\%), power consumption (W), memory usage (GB), and system latency (ms) on the device side. 
 
\noindent\textbf{Resource Consumption on the Server}. We report the latency of the global optimization operation (ms), the overall memory usage (GB), and the system latency (ms) on the server. We also report the network bandwidth demand (Mbps) and the network latency (ms) when multiple users upload map data concurrently. 

\noindent\textbf{Map Quality}.
We adopt Absolute Trajectory Error (\textbf{ATE})(m)~\cite{tum} and Map Reconstruction Error (m)~\cite{beiying} to evaluate map quality. The ATE measures the difference between the ground truth and estimated trajectories.
We calculate ATE with the open-source tool EVO~\cite{evo}. To quantify the Map Reconstruction Error, we measure the distances between the ground truth and map points generated by \systemname and then obtain the root-mean-square error.

\vspace{-4pt}
\subsection{Evaluation of Map Expanding} 
\label{subsec:map-expansion}
\vspace{-2pt}
We first report the map expansion performance of \systemname across the 4 datasets.
The public EuRoC dataset involves 5 users, while the other three settings had a total of 20 users that participate in mapping, each following their own trajectory. Fig.~\ref{fig:traj} shows the trajectories and visualized 3D mapping results of four scenarios. While not intended for human use, these maps provide detailed 3D structural information and can aid in visualization, navigation ~\cite{gao2014jigsaw}, augmented reality~\cite{Maria-conext}, and other application scenarios.

\begin{figure*}[htbp]
\centering
    \begin{minipage}[t]{0.33\linewidth}
        \hspace{-0.1cm}\centering
        \includegraphics[width=2.25in]{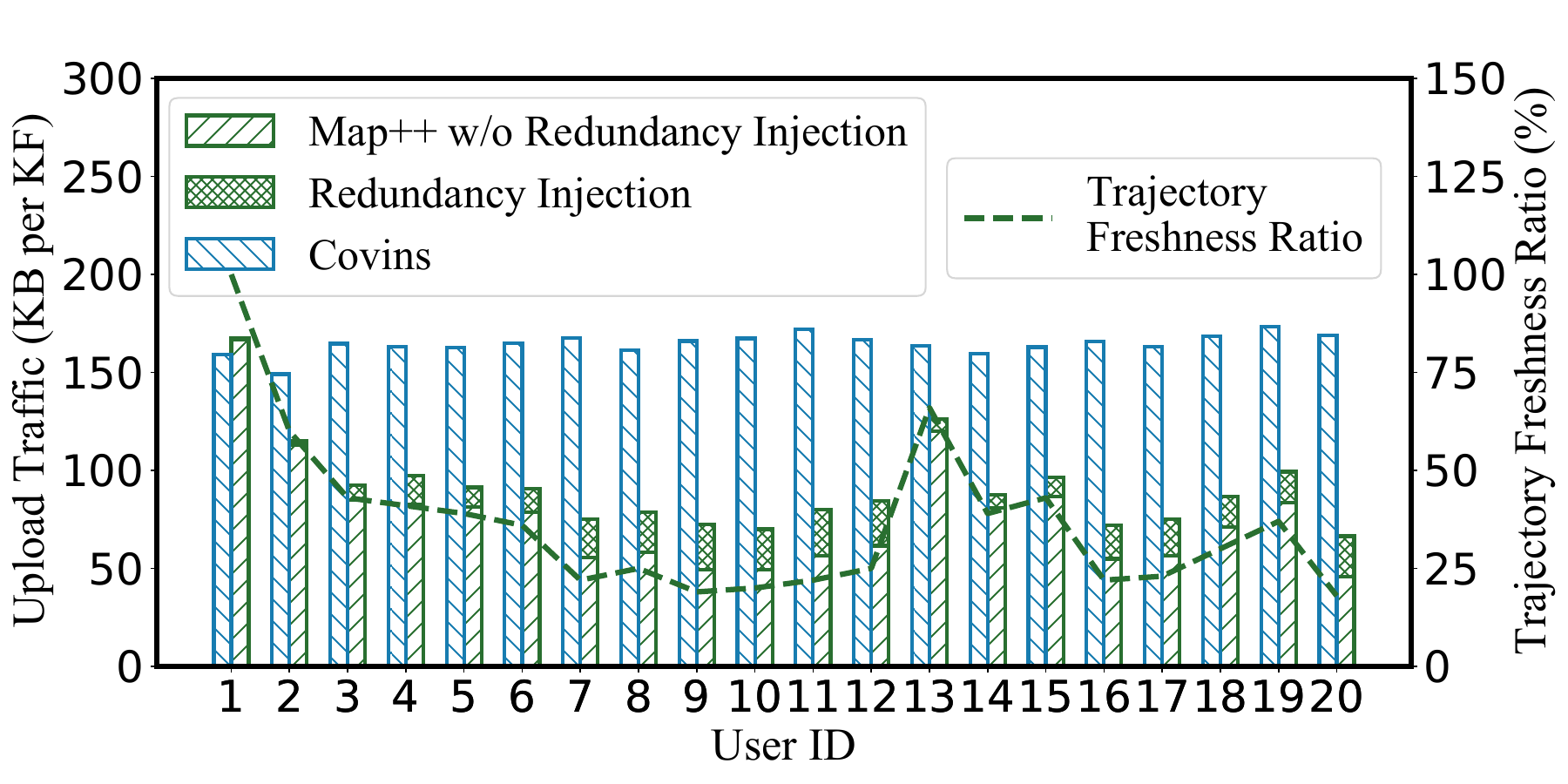}\\
    \captionsetup{width=0.95\linewidth}
    \vspace{-0.3cm}
    \caption{\small\label{fig:traffic1}Upload traffic (left y-axis) and trajectory freshness ratio (right y-axis)  in the \datasetgarage setting.} 
    \end{minipage}%
    \begin{minipage}[t]{0.33\linewidth}
       \hspace{0.15cm}\centering
        \includegraphics[width=2.25in]{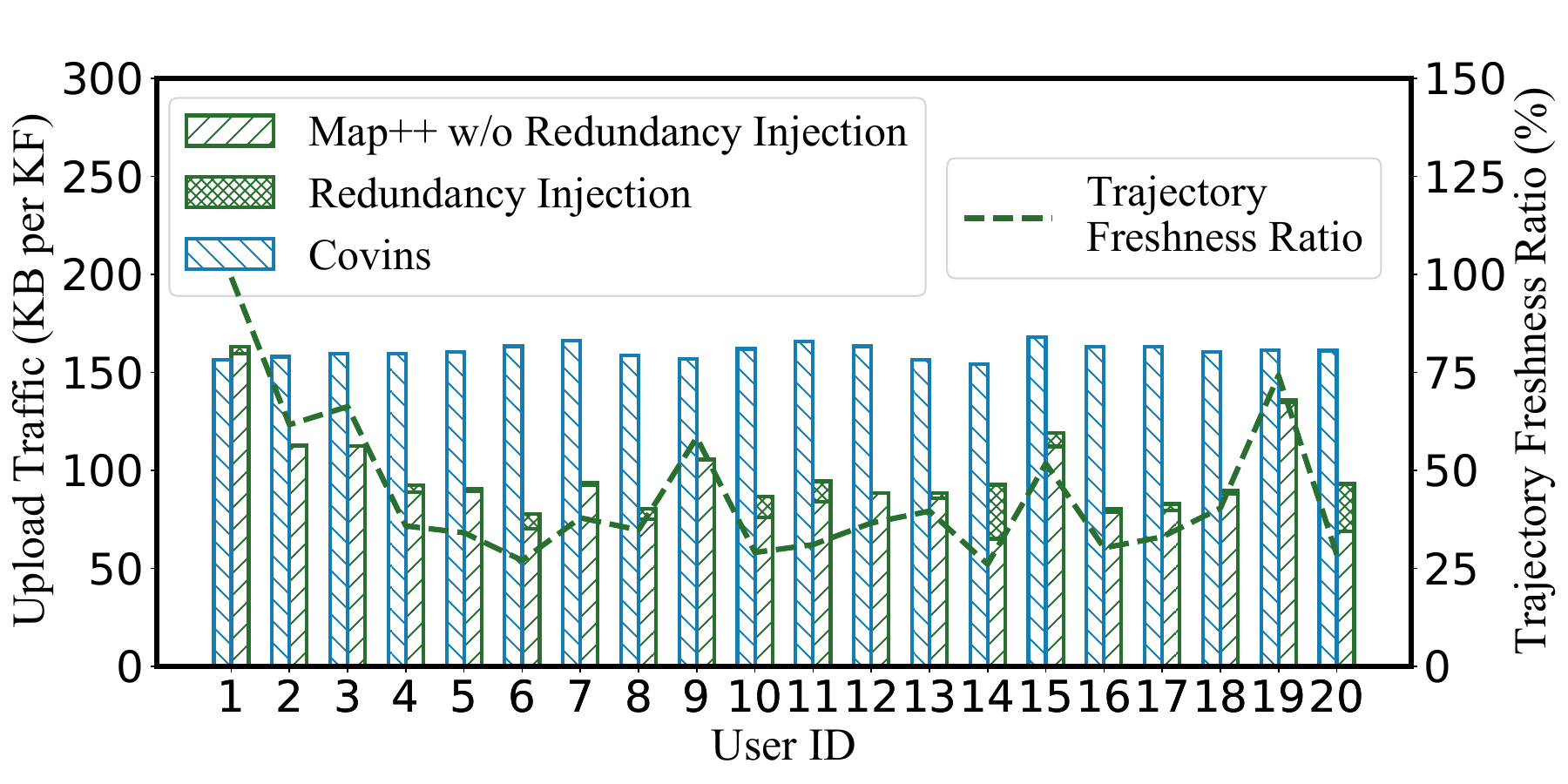}\\
         \vspace{-0.3cm}
        \captionsetup{width=0.95\linewidth} 
        \caption{\small\label{fig:traffic2}Upload traffic (left y-axis) and trajectory freshness ratio (right y-axis) in the \datasetbuilding setting.}
    \end{minipage}%
    \begin{minipage}[t]{0.33\linewidth}
        \centering
        \includegraphics[width=2.25in]{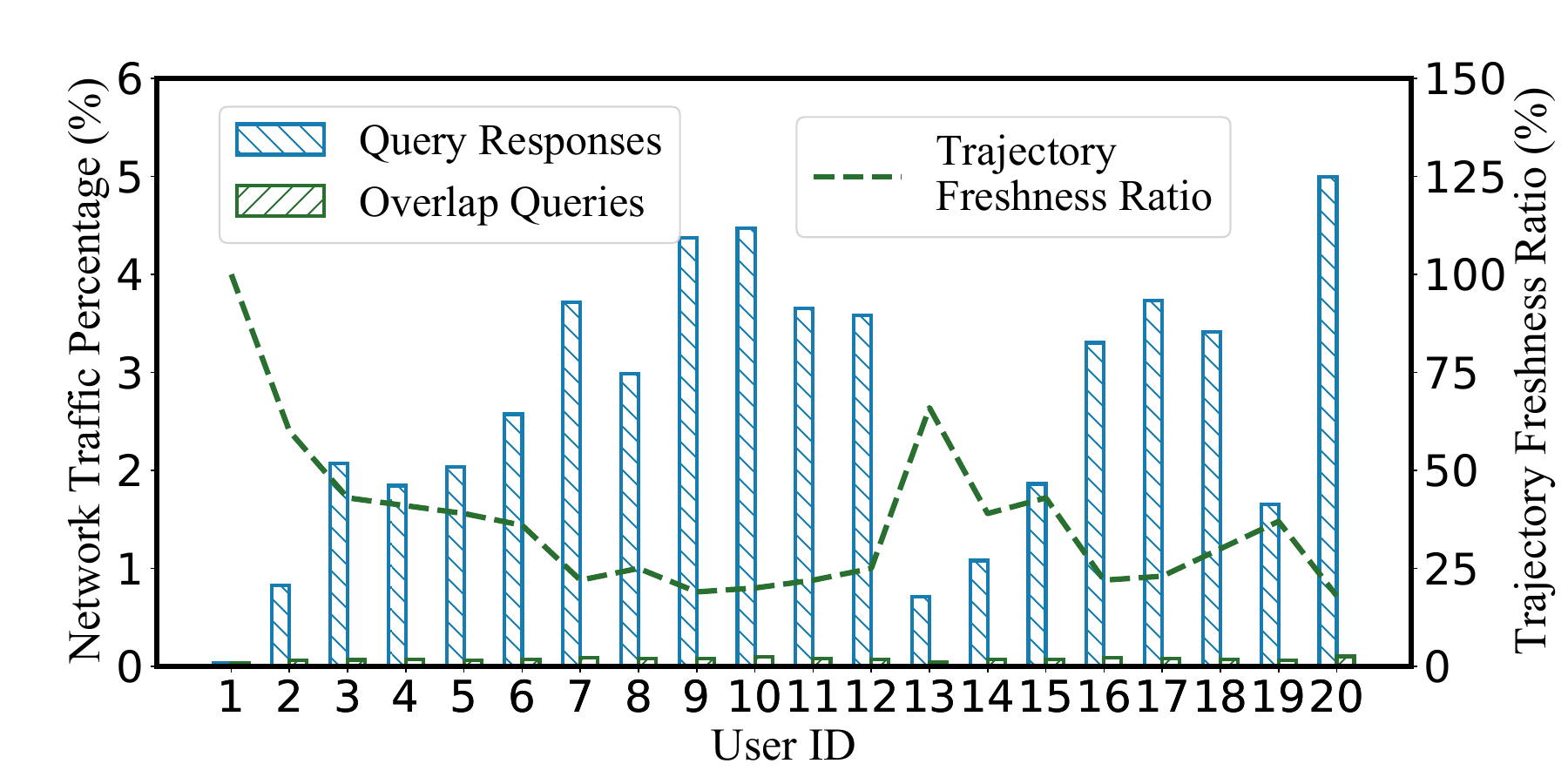}\\
         \vspace{-0.3cm}
        \captionsetup{width=0.95\linewidth}         
        \caption{\small\label{fig:traffic3}Traffic overhead (left y-axis) and trajectory freshness ratio (right y-axis) in the \datasetgarage setting.} 
    \end{minipage}%
    \vspace{-8pt}
\end{figure*}

\subsubsection{Device-Side Resource Consumption} We first report the resource consumption measurements on the device side. 

\noindent\textbf{Upload Data Traffic}. We present the average upload traffic per keyframe (in KB) for each mapping user in the \datasetgarage and \datasetbuilding settings in Fig.~\ref{fig:traffic1} and Fig.~\ref{fig:traffic2}.
In \systemname, we transmit entire keyframes during the cold start. After map merging, the amount of keyframe transmissions declines gradually due to the emergence of overlapping. We also transmit extra redundant data (those map points observed by more keyframes) to improve the global optimization, as discussed in Sec.~\ref{sec:new}.

We observe that Covins incurs very similar upload traffic for each keyframe/user, approximately the size of the original keyframe in both \datasetgarage and \datasetbuilding settings.  
In \systemname, the traffic depends on the freshness of the user trajectory. We use a polyline to represent each user trajectory's freshness ratio (shown on the y-axis on the right side of the figure). To obtain the freshness ratio of a trajectory, we measure the mean overlap ratio ($r_{overlap}$) between the keyframes and the global map. The freshness ratio is $1-r_{overlap}$. A higher freshness ratio means more data is needed to build the global map. Taking \datasetgarage user 10 as an example, our method achieved a significant reduction compared to Covins, saving 54\% of upload traffic, shown in Fig.~\ref{fig:traffic1}. On average, our upload traffic decreases by 46\%. We observe a similar trend for the \datasetbuilding dataset, shown in Fig.~\ref{fig:traffic2}. Due to its open-space nature, this dataset consists of more diverse user paths. Therefore, we observe a slightly less traffic reduction, about 41\% compared to Covins.
On EuRoC dataset, our approach demonstrates noteworthy traffic improvements over the baseline. This leads to a traffic reduction of 60\%, 31\%, 35\%, and 50\% per keyframe for subsequent users, as highlighted in Tab.~\ref{tab:euroc_ATE}.

\begin{table}[b]
	\renewcommand\arraystretch{1}
	\footnotesize \vspace{-12pt}
	\begin{center}
		\scalebox{0.9}[0.9]{
			\setlength\tabcolsep{6pt}
	\begin{tabular}{  |c|c|c|c|c|c|c| }
                \hline 
                \multirow{2}{*}{\makecell{\thead{Item}}} & \multirow{2}{*}{\makecell{\thead{Method}}} & \multicolumn{5}{c|}{Trajectory} \\
			    \cline{3-7} 
		        &   & 1 &2 & 3 & 4  &5  \\		
				\hline  \hline 
				   \multirow{2}*{ ATE (cm) }
                    & \multicolumn{1}{c|}{  Covins }   & \multicolumn{1}{c|}{  1.5  }  & \multicolumn{1}{c|}{   1.7    }    & \multicolumn{1}{c|}{  2.6   }   & \multicolumn{1}{c|}{   4.6   }     & \multicolumn{1}{c|}{  4.6  } \\  
                    \cline{2-7}
    		    & \multicolumn{1}{c|}{  \systemname  }    & \multicolumn{1}{c|}{    2.0 }  & \multicolumn{1}{c|}{    2.4  }    & \multicolumn{1}{c|}{   2.6   }  & \multicolumn{1}{c|}{  4.6   } & \multicolumn{1}{c|}{   7.8  }   \\  
    			 \cline{2-5}
     	      \hline
    				\hline
    	          \makecell{\Gape[0pt][6pt]{\multirow{2}*{ \thead{   Traffic \\ Per KF (KB)  } } }}  
                 & \multicolumn{1}{c|}{  Covins }  & \multicolumn{1}{c|}{  163   }  & \multicolumn{1}{c|}{   168 }    & \multicolumn{1}{c|}{   167  }  & \multicolumn{1}{c|}{  161   } & \multicolumn{1}{c|}{   163  }   \\  
 
    				 \cline{2-7}
      		    & \multicolumn{1}{c|}{    \systemname }  & \multicolumn{1}{c|}{  173  }  & \multicolumn{1}{c|}{   67  }    & \multicolumn{1}{c|}{  116   }  & \multicolumn{1}{c|}{  104  } & \multicolumn{1}{c|}{  81    }   \\  
    				 \cline{2-7}
    				\hline
    			\end{tabular}}   
	\end{center} 
  \caption{\small\label{tab:euroc_ATE} ATE and per-device upload traffic for EuRoC.} \vspace{-10pt}
\end{table}

\noindent\textbf{Discussion: Traffic Overhead of \systemname}. Using \datasetgarage as an illustration, we present the traffic overhead of \systemname in Fig.~\ref{fig:traffic3}, encompassing both overlap query messages (upload traffic) and query response messages (download traffic). The results show that both of them are very small compared to the keyframe size.  
The overlap query size averages around 0.07\% of the keyframe size, while the query response message averages about 2.64\% of the keyframe size.

\subsubsection{Server-Side Resource Consumption} We next report the resource consumption measurements on the server side. 

\begin{figure}[!t]
    \centering
\begin{tabular}{cc}  
\hspace{-0.3cm}
\includegraphics[width=.43\linewidth]{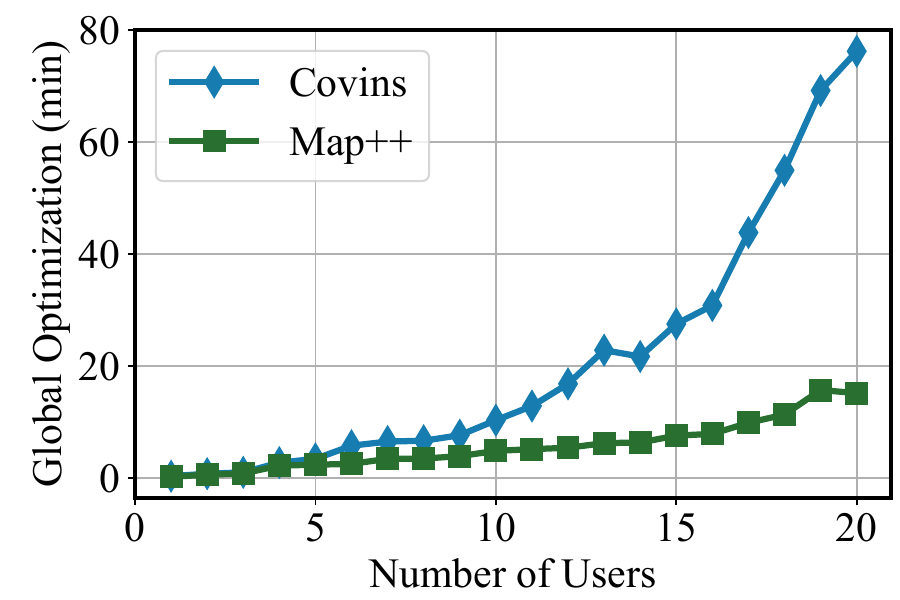} & 
\hspace{-0.3cm}
\includegraphics[width=.43\linewidth]{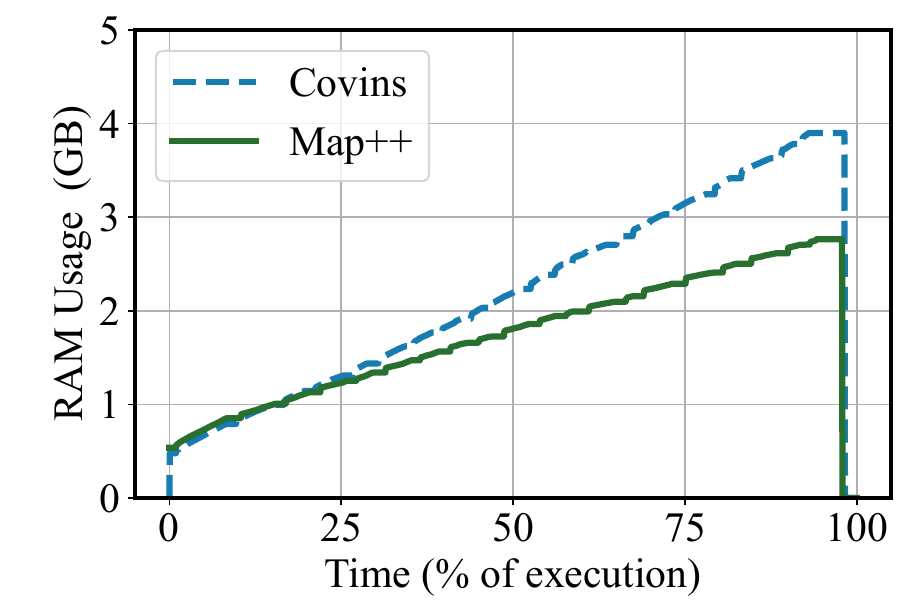} \vspace{-4pt}\\
\small(a) Optimization latency & \small(b) RAM usage \\
\hspace{-0.3cm}
\includegraphics[width=.43\linewidth]{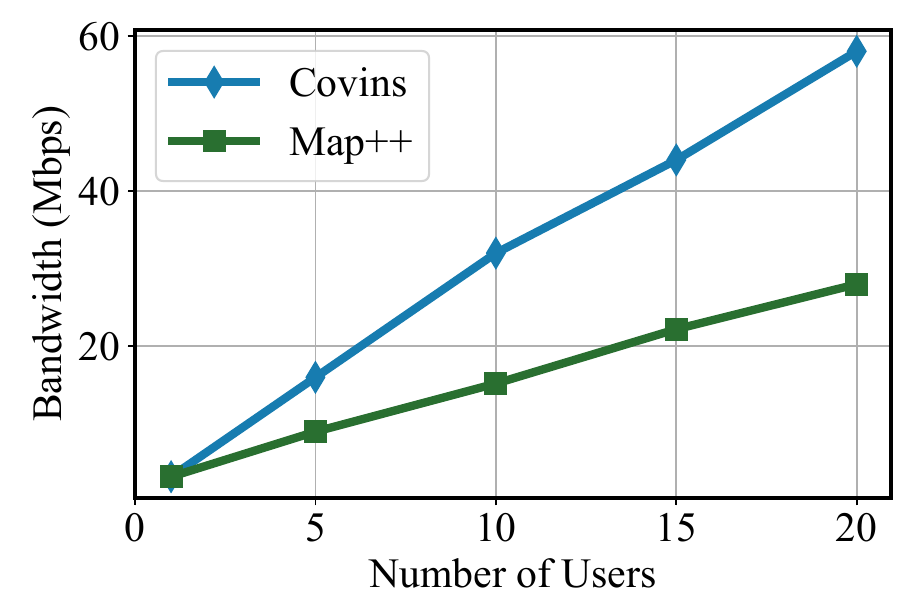} & 
\hspace{-0.3cm}
\includegraphics[width=.43\linewidth]{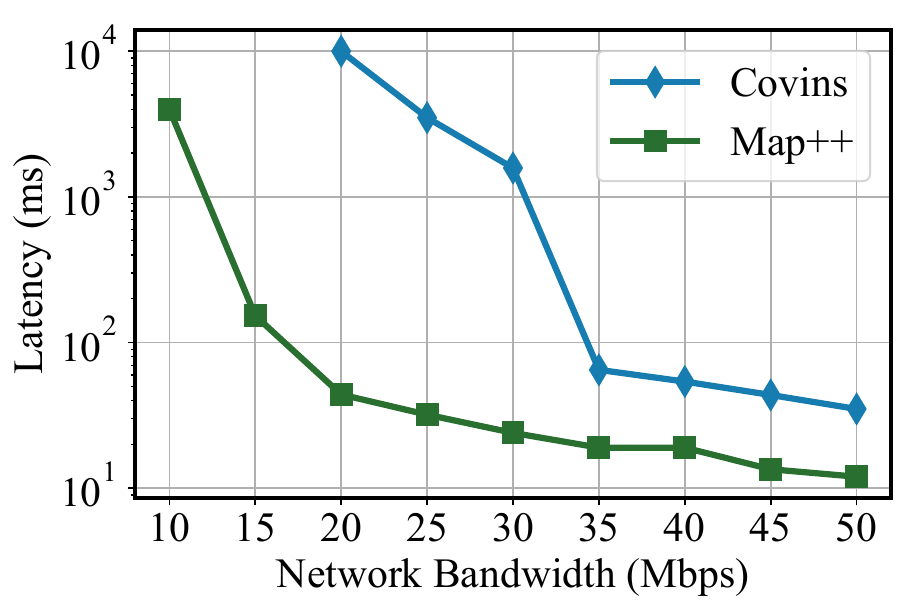}  \vspace{-5pt}\\
\small(c) Bandwidth demand & \small(d) Network latency  \\
\end{tabular}
\vspace{-10pt}
\caption{\small\label{fig:overhead_server} Resource consumption of map expansion on the server side.} 
\vspace{-12pt}
\end{figure}

\begin{table}[b]
\setlength\tabcolsep{1.3pt} 
 \renewcommand\arraystretch{1}
	\footnotesize \vspace{-12pt}
    \begin{center}
	\begin{tabular}{ |c|c|c|c|c|c|c|}
                \hline 
                  \multirow{2}{*}{\makecell{\thead{Method}}}  & \multicolumn{6}{c|}{Latency (ms)} \\
                  \cline{2-7}
                  &  \thead{Overlap  \\  Assessment}      &    \thead{  Redundancy\\ Control}       &   \thead{ KF \\  Upload }   &         \thead{ Map\\  Integration }    &         \thead{  Loc.}   & \thead{  Map \\ Sharing}  \\ 
                 \hline \hline
    	        Covins   &  -  &  -   &  16.5 &    8.4   &  - & -  \\
             	\hline
                   \systemname     &  9.1  &  6.1   & 9.1 & 4.9   &    8.8  &  127.4 \\
             	\hline
    			\end{tabular}
       \end{center} 
  \caption{\small\label{tab:latency_map} System latency. Loc. is the abbreviation of localization.}\vspace{-10pt}
\end{table}

\begin{figure*}[!t]
    \centering
\begin{tabular}{ccc}  
\includegraphics[width=.287\linewidth]{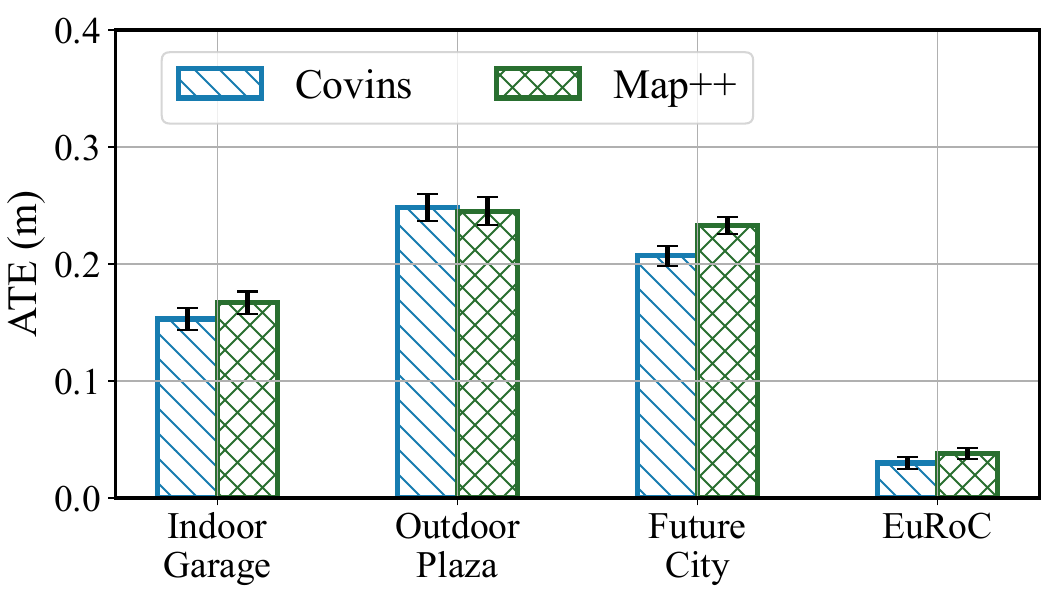} &
\includegraphics[width=.287\linewidth]{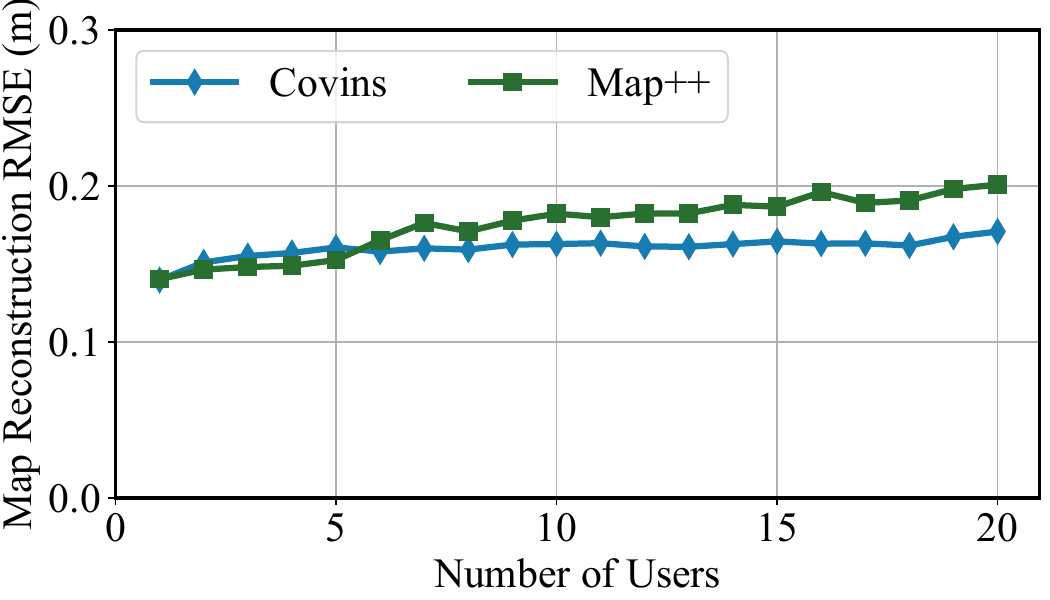}  &
\includegraphics[width=.287\linewidth]{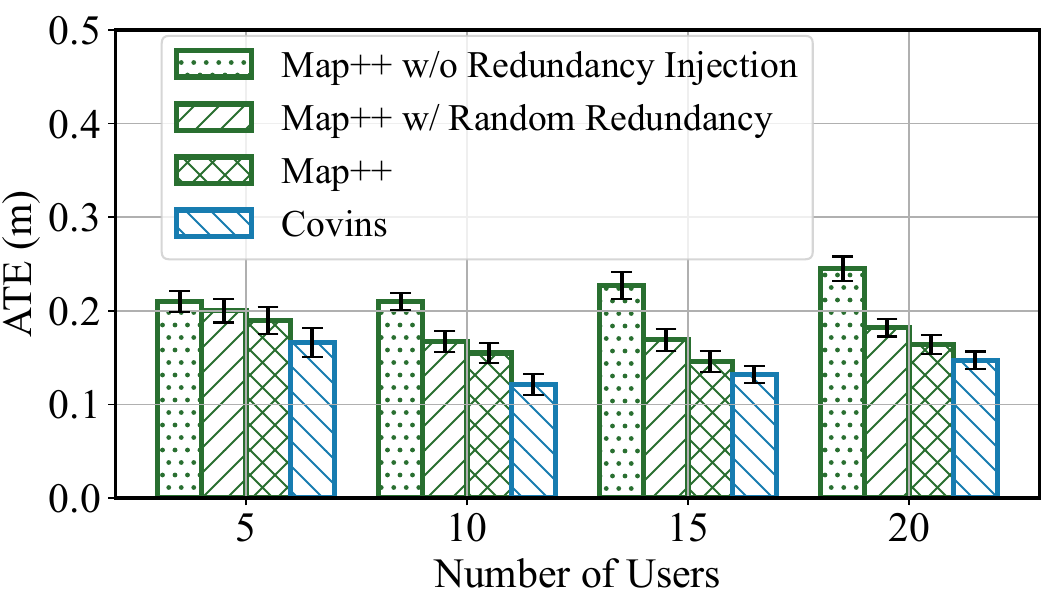} \vspace{-2pt}\\ 
\small(a) Absolute Trajectory Error & \small(b) Map Reconstruction Error & \small(c) ATE for redundancy injection \\ 
\end{tabular}
\vspace{-12pt}
\caption{\small\label{fig:quality} Map quality. (a) Mean ATE for all 4 datasets, (b) Map Reconstruction Error RMSE for \dataset setting, and (c) ATE with and without redundancy injection for the \datasetgarage setting.}
\vspace{-10pt}
\end{figure*} 

\begin{figure*}[!t]
    \centering
\begin{tabular}{ccc}  
\includegraphics[width=.287\linewidth] {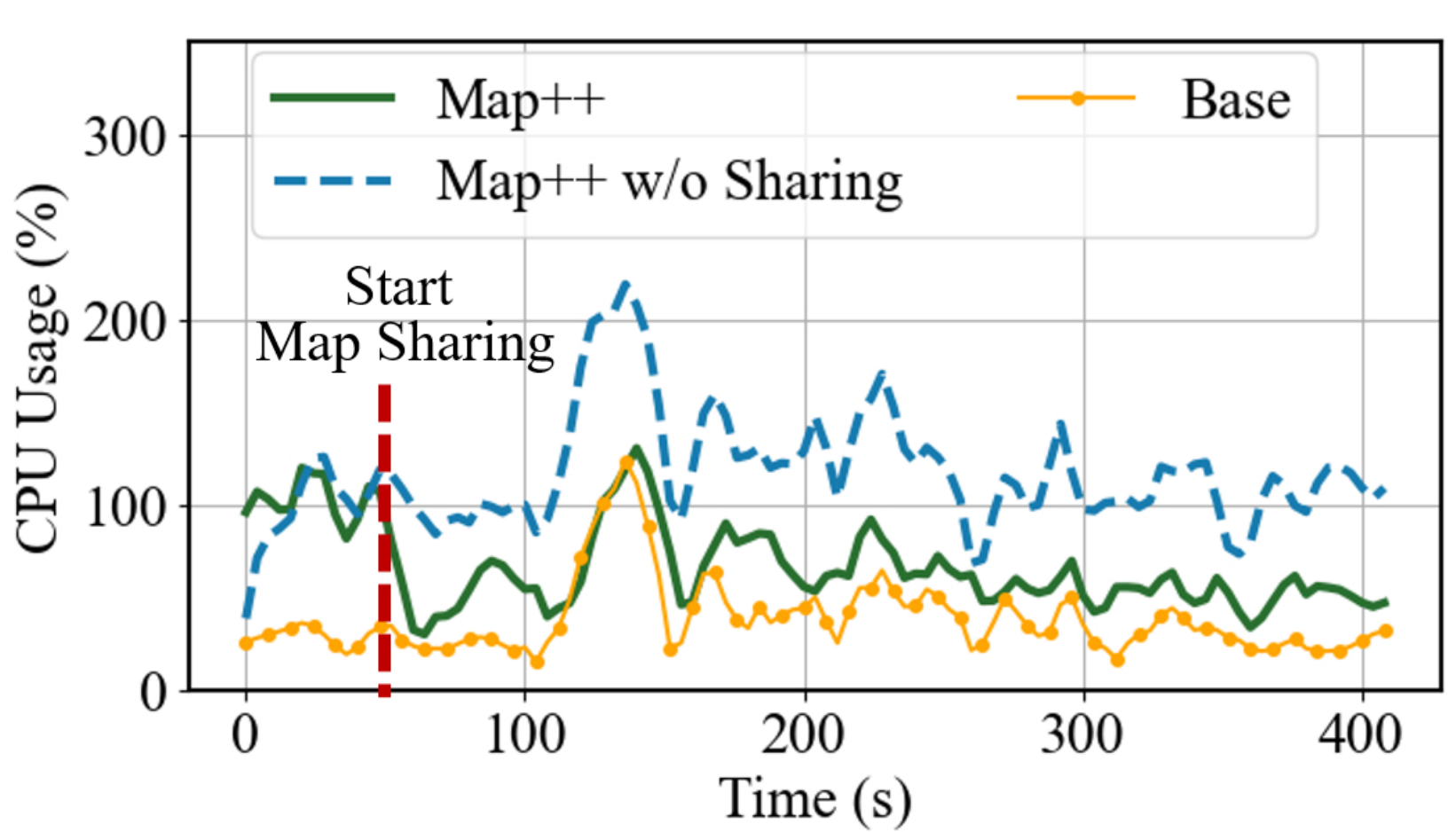} &
 \includegraphics[width=.287\linewidth] {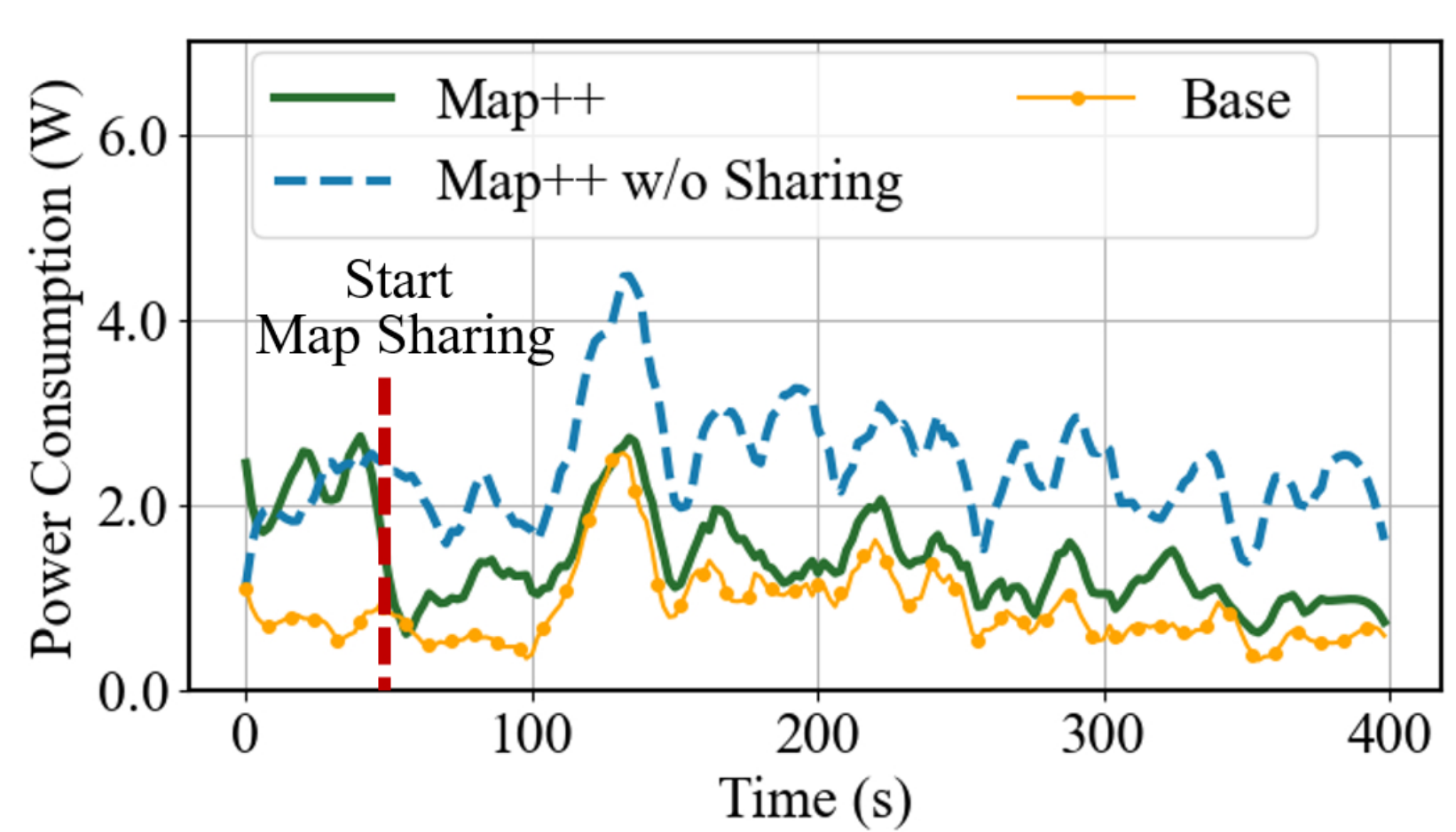}  &
\includegraphics[width=.287\linewidth] {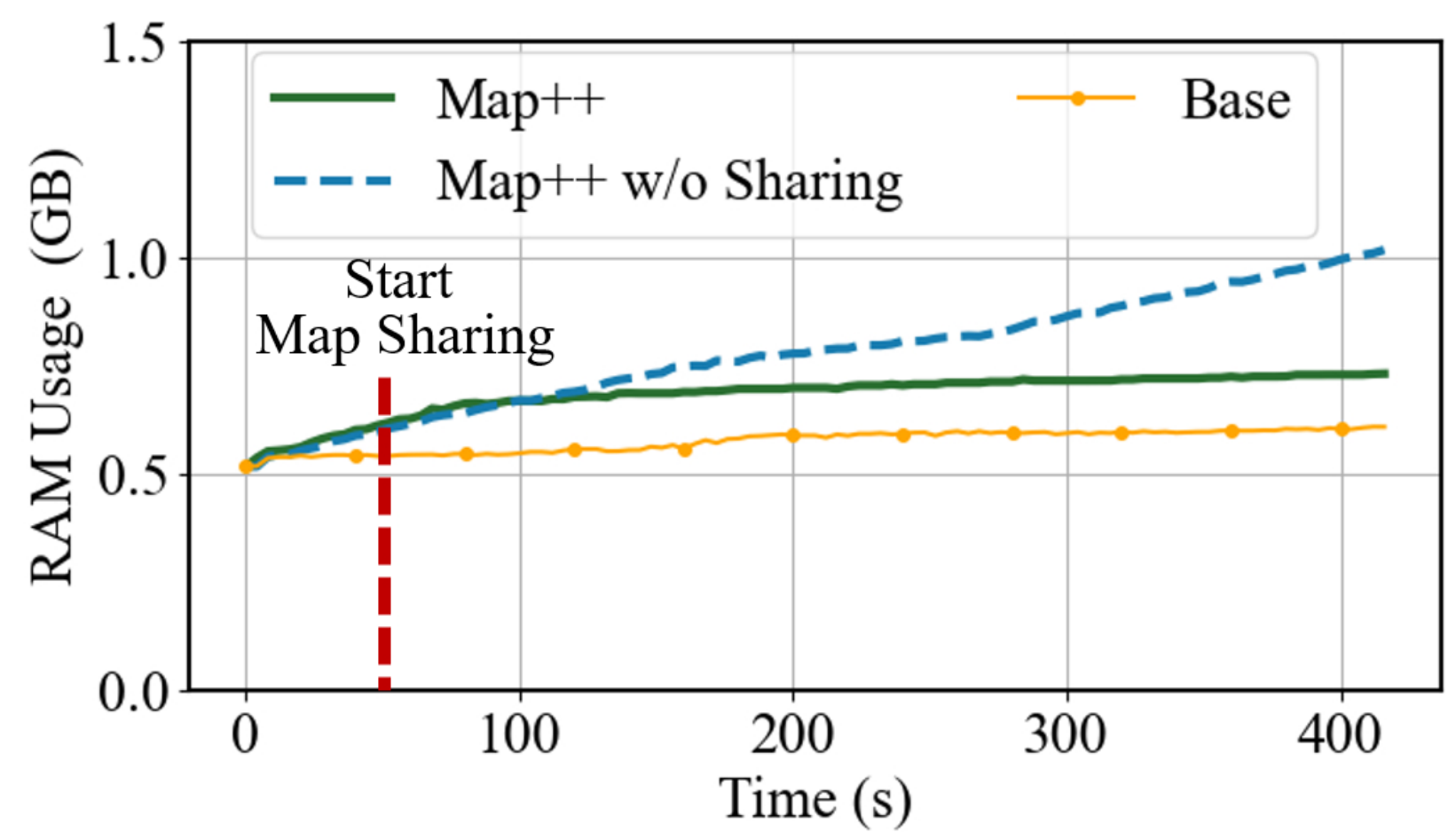} \vspace{-2pt}\\ 
\small(a)   CPU usage on the device& \small(b) Power consumption on the device & \small(c) RAM usage on the device  
\end{tabular}
\vspace{-12pt}
\caption{\small\label{fig:device_cost} Resource consumption in map sharing on the device side. In the experiments, users travel on seen trajectories. Each user first registers with the server, and then starts the map sharing phase. The start time of the map sharing is marked.} 
\vspace{-10pt}
\end{figure*} 

\noindent\textbf{Global Optimization Latency.}
In both Covins and \systemname, we trigger a global optimization when a user's session concludes. This global optimization process optimizes all the map data, ensuring the overall accuracy and consistency of the global map. Notably, \systemname exhibits a significant reduction in global optimization latency as shown in Fig.~\ref{fig:overhead_server}(a). Compared with \systemname, Covins shows a steeper increase, consuming 76 minutes when we have 20 users. This shows that \systemname exhibits much better scalability in terms of the number of users due to much-reduced processing overhead. 
 
\noindent\textbf{Server RAM Usage}.
We measured the RAM usage for Covins and \systemname, and show the results in Fig.~\ref{fig:overhead_server}(b). Our method shows a reduction of approximately 30\% in RAM usage on the server when we have 20 users. As the number of users increases, this gap will continue to widen as more redundancy exists among them while Covins is completely unaware of this redundancy.

\noindent\textbf{Server Network Bandwidth}. We report the bandwidth demand on the server when multiple users upload data concurrently in Fig.~\ref{fig:overhead_server}(c). We observe that \systemname reduces the server bandwidth requirements by approximately 43\%, 47\%, 49\%, and 50\%, with 5, 10, 15, and 20 users. Therefore, with the same bandwidth resources, \systemname can scale up to approximately 2 $\times$ more users than its baseline.

\noindent\textbf{Network Latency}. In Fig.~\ref{fig:overhead_server}(d), we report the network transmission latency (in log scale) under different bandwidth constraints, with 10 users uploading map data concurrently. Results show that \systemname has a much shorter latency compared to Covins. Specifically, when the total bandwidth is below 20Mbps, Covins cannot work properly.  Between 20Mbps and 30Mbps, \systemname incurs a much shorter latency than Covins, i.e., the reduction is 98\% at 30Mbps.
When the bandwidth increases to 35Mbps, both systems work more gracefully, with \systemname reducing the latency by 70\% approximately. This set of results shows that \systemname can handle extreme situations with severe bandwidth bottlenecks much better than Covins.

\subsubsection{Map expansion latency.}
We present the latency of map expansion in Tab.~\ref{tab:latency_map}.  After a keyframe is generated, it undergoes a time-consuming local optimization, with an average duration of 400ms as reported in Fig.~\ref{fig:local}. Subsequently, \systemname utilizes overlap assessment and redundancy control to balance map quality and efficiency, with latencies of 9.1ms and 6.1ms. Due to the redundancy removal, \systemname requires less time for keyframe upload and map integration compared to Covins, reducing the keyframe upload time from 16.5ms to 9.1ms and the map integration time from 8.4ms to 4.9ms.

\subsubsection{Map Quality.} 
We next report the map quality results.

\noindent\textbf{Absolute Trajectory Error.}  In Fig.~\ref{fig:quality}(a),  we present the mean Absolute trajectory error (ATE) for all four settings. In comparison to Covins, \systemname has slightly larger errors: an increase of 0.015m in the \datasetgarage setting, an increase of 0.026m in the \dataset setting, and an increase of 0.009m in the EuRoC setting.  In the \datasetbuilding setting, the two perform similarly. When comparing the result of \datasetgarage and \datasetbuilding settings, we observe larger errors in the latter. Images captured in strong outdoor lighting conditions are prone to overexposure. Accordingly, the features extracted from these images become unstable, resulting in larger errors. 

\noindent\textbf{Map Reconstruction Error}. 
The simulated \dataset provides us with trajectory ground truth, as well as a 3D mesh ground truth.  Therefore, we can report the map reconstruction error on this dataset in Fig.~\ref{fig:quality}(b). When we have more users, the map reconstruction error gradually increases. 
When there were 20 users, compared to Covins, \systemname shows an increase of 0.03m. 

\noindent\textbf{Discussion: Data Redundancy}. 
Indeed, having more redundant data often leads to more effective global optimization and higher mapping accuracy. Our results also confirm this. However, given the considerable traffic volume reduction as well as other resource consumption reduction in \systemname, we believe the minor compromises in mapping quality, as shown above, are quite acceptable. Moreover, our resource conservation will be more pronounced as the number of users increases. We also took a close look at the data collected from \datasetgarage to show the effectiveness of our redundancy injection method in Fig.~\ref{fig:quality}(c). 
Specifically, we compare our proposed redundancy injection mechanism with random redundancy injection.
We observe that both methods can reduce ATE for \systemname. However, by uploading those map points that are observed more often, \systemname can bring a more competitive performance.
For example, when the system has 20 users, the proposed redundancy control technique and the random redundancy injection can reduce the ATE from 0.245m to 0.164m and 0.182m, respectively.

\begin{table}[b]
 \renewcommand\arraystretch{1}
 \arrayrulecolor{black} 
	\footnotesize \vspace{-15pt}
	\begin{center}
	\begin{tabular}{|c|c|c|c|c|}
                        \hline 
                \multirow{2}{*}{\makecell{\thead{Method}}}   & \multicolumn{2}{c|}{ \thead{ATE (m)}} & \multicolumn{2}{c|}{ Map sharing traffic per KF (KB) }   \\  
			    \cline{2-5} 
		        &  Garage & Plaza & Garage  &Plaza   \\		
				\hline  \hline 
    	 
        w/o sharing      &   0.175   &  0.308     &   -  &   -   \\  
        \hline 
        sharing      &   0.128   &  0.280     &   25  &   24   \\  
    			    \cline{2-5}
    				\hline
    			\end{tabular}
	\end{center}
  \caption{\small\label{tab:localization} Map sharing leads to better localization accuracy.}
\end{table}

\vspace{-4pt}
\subsection{Evaluation of Map Sharing} 
\label{subsec:map-sharing} 
\vspace{-2pt}
In our experiments, we employ the real-world \datasetgarage and \datasetbuilding datasets,  to assess the performance of map sharing.  In each setting, we had 15 localization users whose trajectories have been mapped before.   
\systemname engages map sharing to allow users to reuse existing global map on the server, thereby reducing their own resource consumption. As \systemname is deployed for a long-term service, we expect most of the users to fall within this category and can enjoy the map service at a low cost. Since Covins does not support map sharing at all,  we compare two variations of \systemname in this part: (1) with map sharing (oversharing factor $\alpha=1.3$) and (2) without map sharing.  In the latter system setting, for a seen pose, the server does not share the global map with the device. The device performs local optimization with its local data without uploading the data for mapping purposes.

\begin{figure}[!t]
    \centering
\begin{tabular}{cc}  
\hspace{-0.2cm}
\includegraphics[width=.8\linewidth]{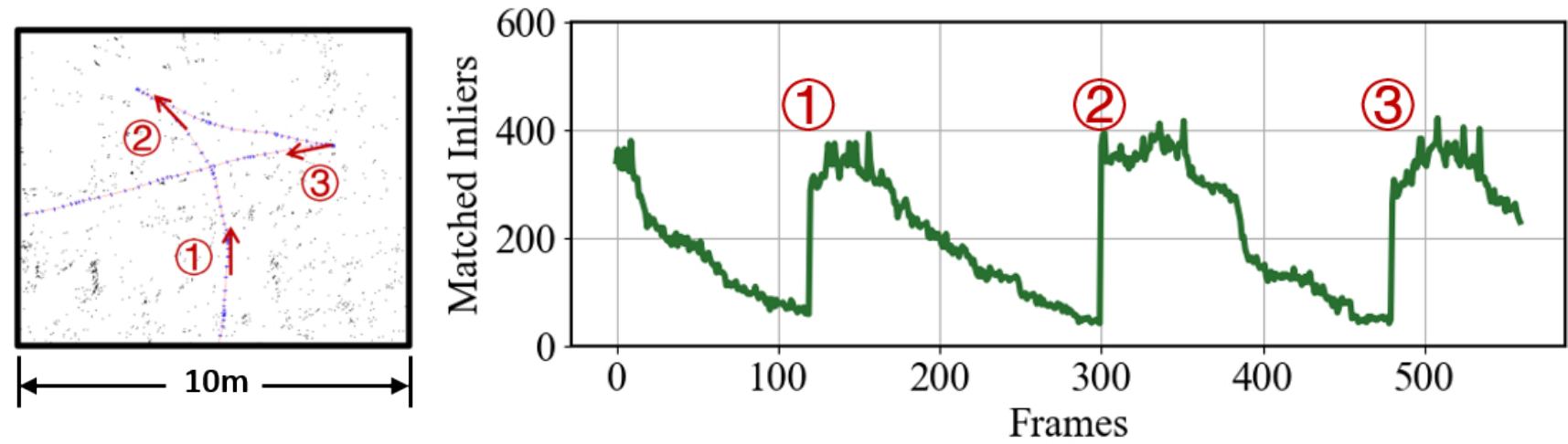}  \vspace{-6pt}\\
\small(a)  User with $\alpha=1$  \\
\hspace{-0.2cm} 
\includegraphics[width=.8\linewidth]{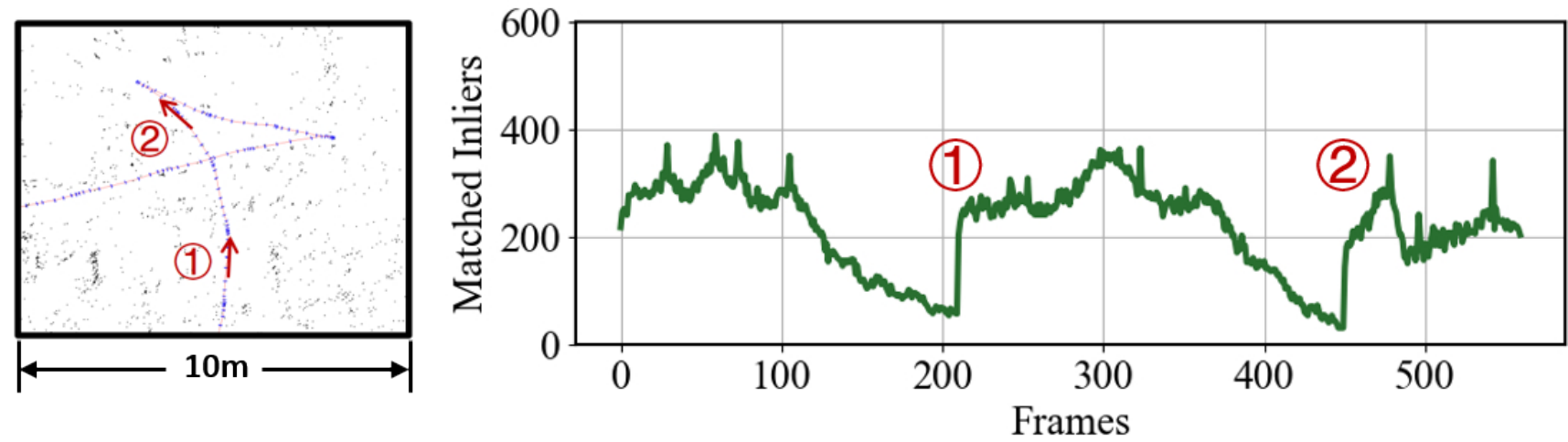} \vspace{-6pt}\\
\small(b)  User with $\alpha=1.3$  \\
\end{tabular}
\vspace{-14pt}
\caption{\small\label{fig:loc_failure} (a) The user with $\alpha=1$ experiences  3 map requests over a 22-meter path. (b) The user with $\alpha=1.3$ experiences 2 map requests for the same path.  The plots on the left show the user's trajectory and local map, and the plots on the right show the number of matched map points for each frame.}
\vspace{-10pt}
\end{figure} 

\subsubsection{Device-Side Resource Consumption} We first report the resource consumption measurements on the device side in Figs.~\ref{fig:device_cost}(a)-(c). Here, we also measure the resource consumption for base SLAM functions -- i.e., device-side \systemname without the localization module  -- including ROS message subscription, ORB feature extraction, and synchronization of camera and IMU. In this way, we can better understand the impact of map sharing on device localization.

\noindent\textbf{Device CPU Usage}. To demonstrate the effectiveness of map sharing, we compare the CPU usage of user devices with and without map sharing. 
Fig.~\ref{fig:device_cost}(a) shows the CPU utilization during a user's runtime. In the initial 50 seconds (earlier than the red line), the user needs to register with the server while running the same routine as in map expanding phase. After the registration phase, the server identifies the user is on a seen path and starts map sharing. The user utilizes shared maps for localization without running local optimization. As shown in Fig.~\ref{fig:device_cost}(a), the total CPU usage can be decreased by 48\% on average by skipping the local optimization function. If we take a close look at the localization module by extracting the base usage curve from the two \systemname curves, we observe that the localization module's CPU usage is cut down by 72\% due to map sharing.  Please note that in  Fig.~\ref{fig:device_cost} we use image resolution of 1280x720. The observed trend also holds for other camera parameter settings. 

\noindent\textbf{Device Power Consumption}. 
We report the power consumption of the device in Fig.~\ref{fig:device_cost}(b). For the measurement duration of 410 seconds, despite fluctuating, map sharing can consistently lower the total power consumption by 47\%. If we focus on map sharing and extract the base usage curve, the average power consumption reduction is 75\%.

\noindent\textbf{Device RAM Usage}. 
Map sharing can not only reduce CPU usage but also decrease RAM memory usage on the device, as illustrated in Fig.~\ref{fig:device_cost}(c). 
Without map sharing, the RAM usage shows a much faster increase rate over time.  In fact, as the user trajectory continues to extend, the benefit of map sharing will become much more pronounced.

\subsubsection{Map sharing latency.} 

We also report the latency of map sharing in Tab.~\ref{tab:latency_map}. Map sharing serves multiple frames in one execution, consuming 127.4ms per execution. The device requires a shared map only when the user experiences a localization failure. For all other times, the user employs the shared map as its local map. Benefiting from map sharing, \systemname can localize in 8.8ms without the need for time-consuming local optimization. 

\subsubsection{Device Localization Accuracy}
We next present the device localization accuracy ATE in \datasetgarage and \datasetbuilding settings.
Map sharing can lead to more accurate localization because the shared map from the server has gone through global optimization and is thus more accurate.  The mean ATE with map sharing is $0.128 m$ and $0.280 m$  for the two data sets, as shown in Tab.~\ref{tab:localization}, outperforming the system without map sharing by $0.047m$ and $0.028m$.    
Please note that our system can provide users with a 6-DoF pose and a detailed 3D map, which is much richer than traditional indoor WiFi localization systems such as~\cite{ayyalasomayajula2020deep,chen2019m,xie2019md}.

\noindent\textbf{Discussion: Proactive Map Sharing.} We next illustrate the advantages of proactive map sharing. In Fig.\ref{fig:loc_failure}(a), we set the oversharing factor $\alpha = 1$, wherein the user made three requests for shared maps given a 22-meter path.  We observe that the number of matched map points reaches the peak when receiving a shared map and decreases gradually until the user needs to request a new shared map.  In Fig.~\ref{fig:loc_failure}(b), where $\alpha = 1.3$, only two map requests are issued. Each requested map can support a longer period of 240 frames. This example clearly demonstrates the advantage of proactive map sharing, which is one of the main features of \systemname.

\begin{figure}[t]
\begin{tabular}{ccc}
\includegraphics[width=.28\linewidth] {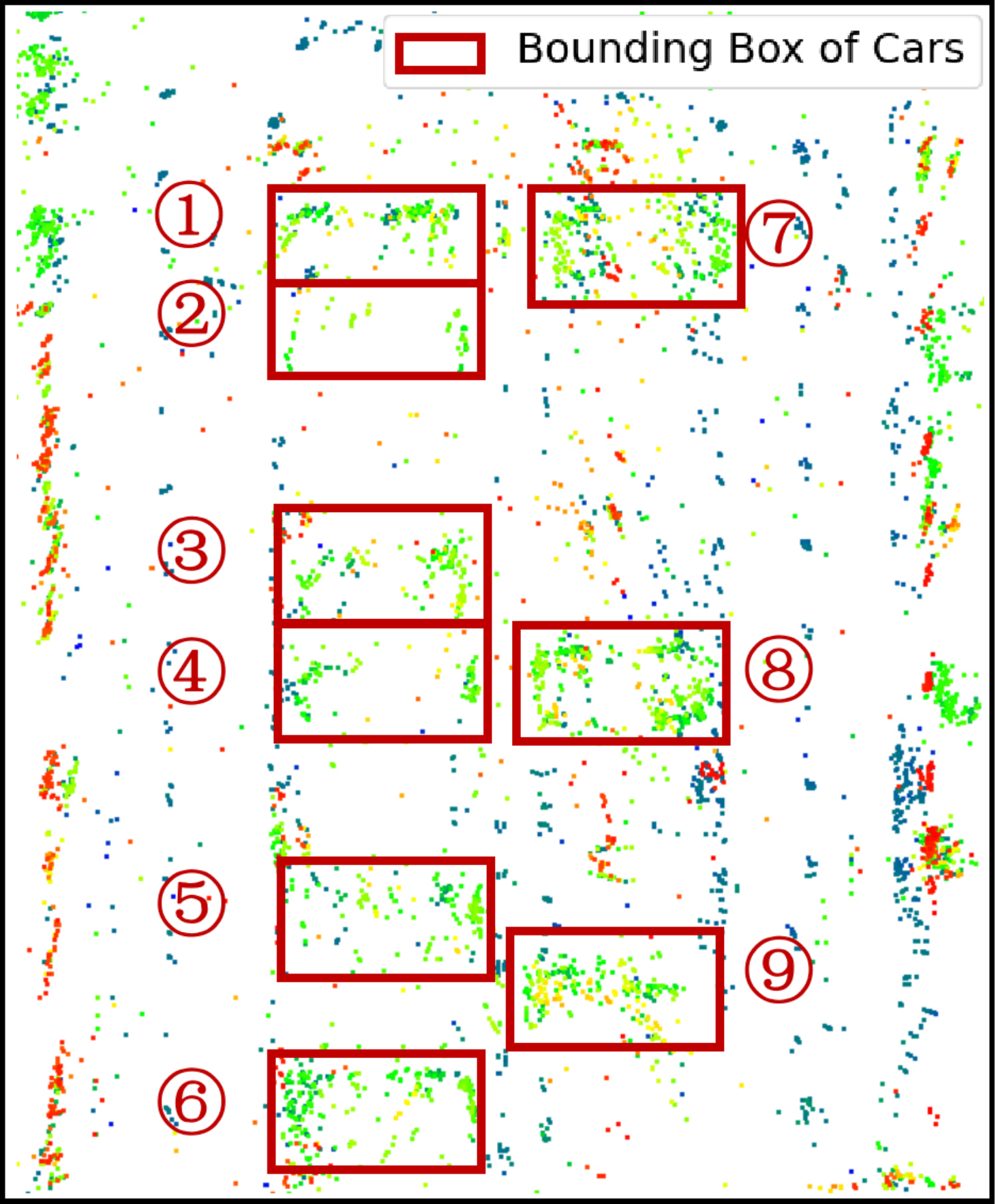} &  \hspace{-0.1cm}
\includegraphics[width=.28\linewidth] {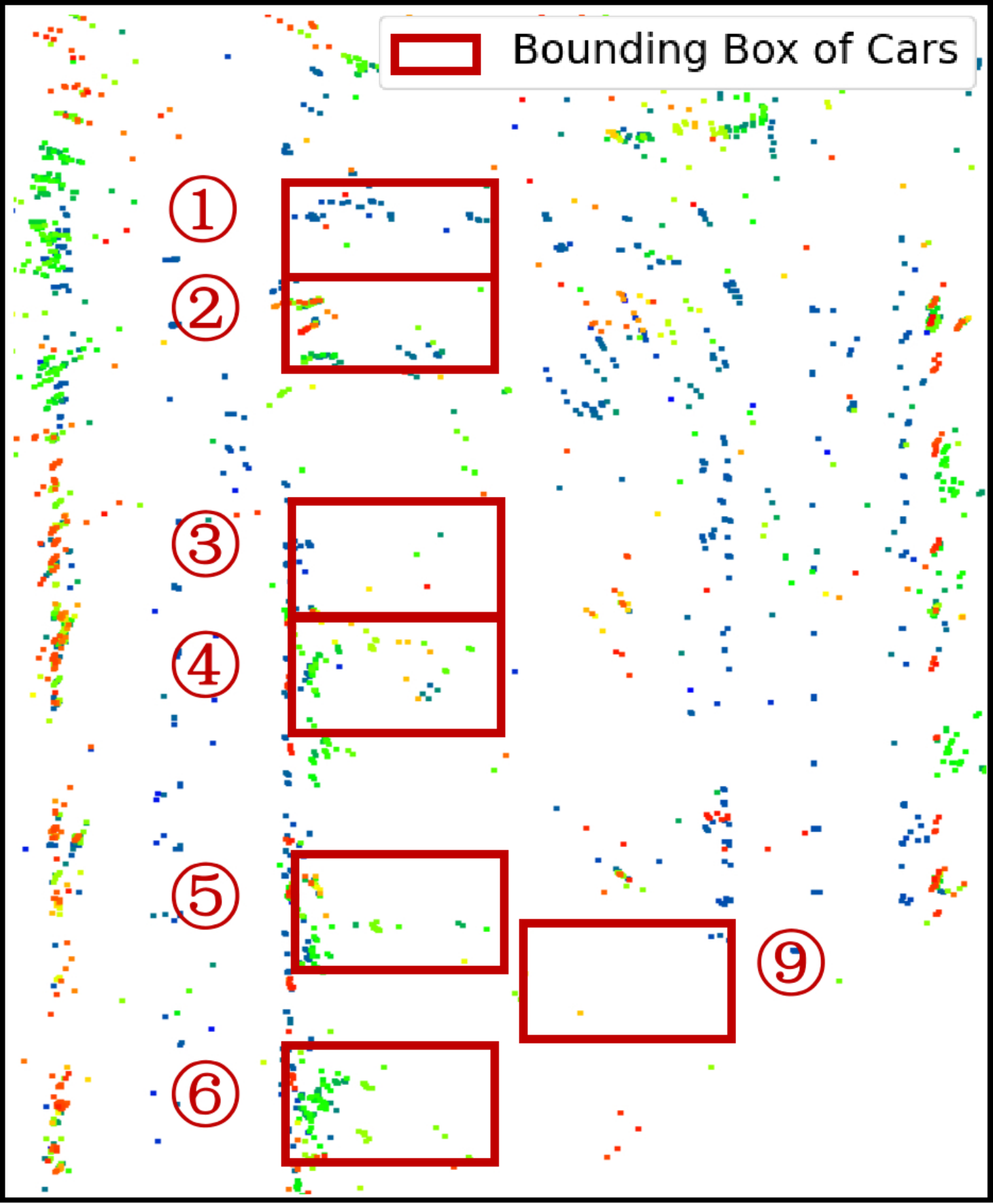} &
\hspace{-0.1cm}
\includegraphics[width=.28\linewidth] {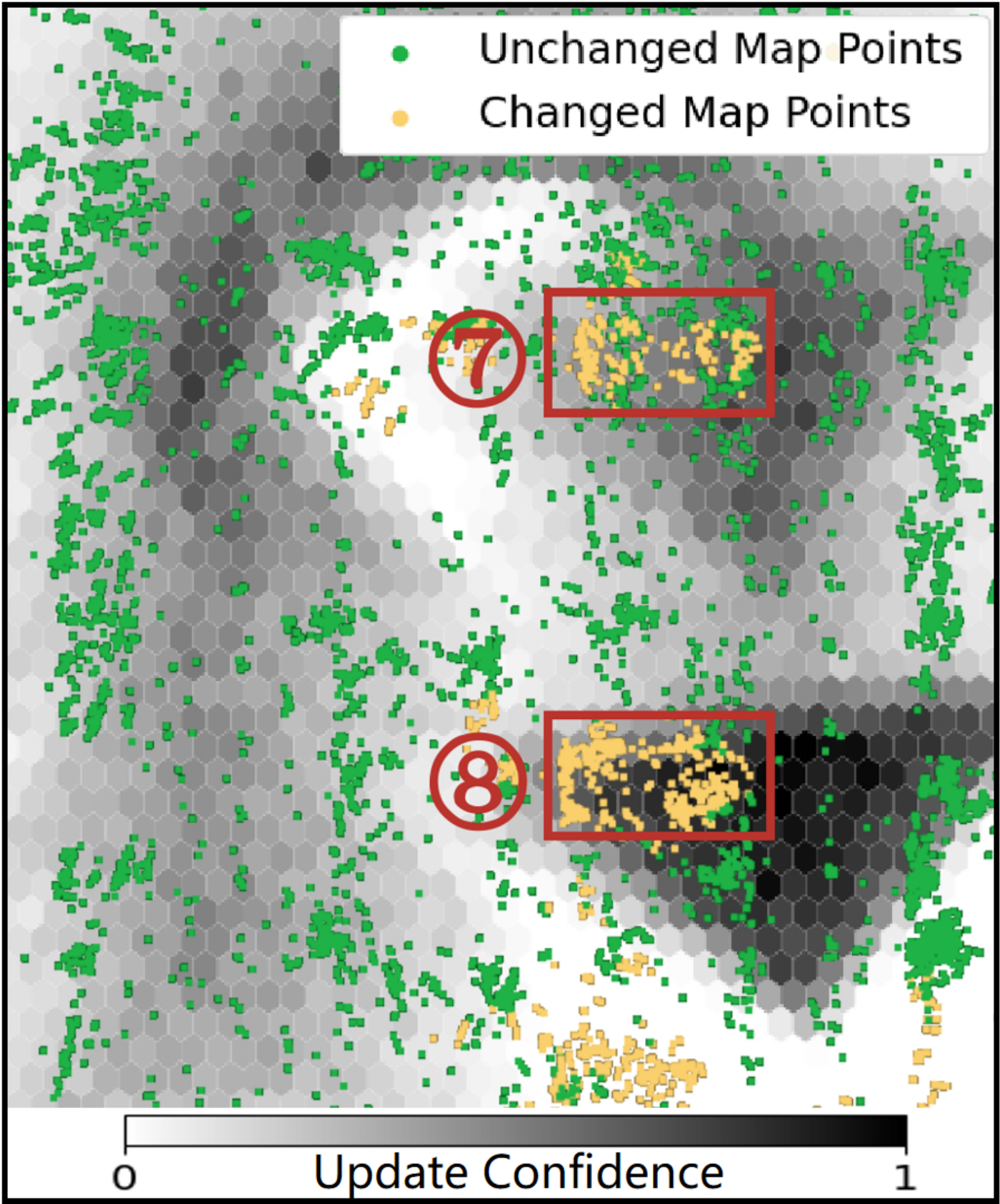} \vspace{-2pt}\\
\small \hspace{0.0cm}(a) & \small \hspace{-0.1cm}(b) & \small \hspace{-0.1cm}(c) \\
\end{tabular}
\caption{\small\label{fig:update_detection} Outdated global map can be detected. (a) shows an outdated global map, (b) shows the new map generated from a user's recent keyframe that captures environment changes, and (c) shows that our system can detect the changes.}
\end{figure} 

\subsubsection{Global Map Updating}  
We also show the feasibility of detecting outdated map data using the \datasetgarage dataset. As illustrated in Fig.~\ref{fig:update_detection}(a),  we show the bounding boxes for the nine cars numbered $1$ through $9$. The global map in Fig.~\ref{fig:update_detection}(a) was built by the first four users. Then, two cars, 7 and 8, left the garage before the fifth user joined. The fifth user joined the system and utilized the existing maps.
The fifth user then built the local map, shown in Fig.~\ref{fig:update_detection}(b) when he found that the environment had changed. 
The final change detection results are shown in Fig.~\ref{fig:update_detection}(c). The points to be updated are colored orange. 
Considering the confidence levels, we identify two actual scene changes marked in red. Please note that our study focuses on detecting the scene changes, not on how to perform the update operation.

\section{Related Work}\label{sec:related} 
\noindent\textbf {Simultaneous Localization and Mapping.} When exploring an unknown space, the most effective strategy is to construct a map while simultaneously locating on it. Maps can be constructed from diverse sources, such as images~\cite{MonoSLAM, engel2017direct,orbslam3}, LiDAR point clouds~\cite{loam}, Wi-Fi signals~\cite{ferris2007wifi,sigcomm-rf,jingaoxu-wifi-dof}, electromagnetic fields~\cite{slam-fingerprint}, acoustics~\cite{sensys-acustic-slam}, mmWave~\cite{mmwave,mmwave-pose}, etc. Despite being based on ORB-SLAM3, \systemname can be extended to various point-based SLAM systems.   
  
\noindent\textbf {Multi-user Localization and Mapping.} Harnessing collective efforts from multiple users is pivotal for convenient, low-cost, and continuous mapping. Jigsaw~\cite{gao2014jigsaw} collects crowdsensed images from mobile users for indoor floor plan reconstruction. SLAM-share~\cite{SLAMshare} offloads most of the SLAM computations and transmits raw camera data to the server. It performs tracking and mapping on the server to build a shared global map. Additionally, they aim to speed up tracking and map merging by leveraging GPU acceleration and shared memory. Conversely, CCM-SLAM~\cite{CCM-SLAM}, CVI-SLAM~\cite{CVI-SLAM}, and Covins~\cite{covins} propose to offload resource-intensive tasks. Meanwhile, they ensure each user's autonomy by executing tracking on the user device. Besides, Covins~\cite{covins, redundancy} selects and removes redundant keyframes to reduce the optimization time.
Moreover, Pair-Navi~\cite{jingaoxu-infocom,jingaoxu-tsn} explores Peer-to-Peer user cooperation by reusing a previous traveler’s trajectory for future user's pose estimation. 
Different from previous vanilla SLAM architectures, \systemname 
can execute pose estimation on the user device to maintain autonomy, it transmits a controllable keyframe to the server. \systemname combines maps from multiple users to create a global map on the server, which can be shared with multiple followers for lightweight localization.
 
\noindent\textbf {Resource Constrained Map Data Transmission.}  
In the context of edge computing, transmitting map data can become a bottleneck. AdaptSLAM~\cite{adaptslam} proposes a theoretically grounded method to assess the uncertainty of each keyframe and transmit only the most critical ones.  CarMap~\cite{carmap} and SwarmMap~\cite{xu2022swarmmap} focus on map updating. CarMap presents a lean representation of 3D map along with a GPS-based feature-matching method for fast feature comparison. By comparing the newly detected features with features on the base map, CarMap identifies differences and updates all the base map copies, whether they are onboard or on the cloud server. 
SwarmMap proposes to execute daemons on both the device and server side for map synchronization, monitoring the function calls, and transmitting the operation logs.
In contrast to compressing the map data directly, \systemname utilizes lightweight metadata to interact with the global map and identify redundant areas. \systemname partitions the keyframe and transmits only the necessary data to the server.

\section{Conclusions} 
\label{sec:conc} 
We have presented the design and implementation of \systemname, a participatory visual SLAM system.
Incorporating efficient redundancy detection and removal techniques, \systemname ensures a careful balance between map quality and efficiency. Furthermore, the proposed framework facilitates long-term map maintenance and enables map sharing among users. \systemname offers a practical approach to mapping, with the potential to revolutionize our abilities in building and maintaining 3D maps in spots close to where live and work (such as parking garages) or completely unknown spaces (such as deep space exploration), both too costly for commercial map vendors.  
Moving forward, we will continue to investigate the numerous issues in deploying such a system, such as dealing with low-quality sensor data and avoiding map corruption.

\begin{acks}
This work is supported in part by the National Natural Science Foundation of China (No. 62332016). 
Yanyong Zhang is the corresponding author.
\end{acks}
 
\bibliographystyle{ACM-Reference-Format}
\bibliography{0_main}
 \end{sloppypar}
 
\end{document}